\documentclass{article}

\usepackage{PRIMEarxiv}

\usepackage[utf8]{inputenc} % allow utf-8 input
\usepackage[T1]{fontenc}    % use 8-bit T1 fonts
\usepackage{hyperref}       % hyperlinks
\usepackage{url}            % simple URL typesetting
\usepackage{booktabs}       % professional-quality tables
\usepackage{amsfonts}       % blackboard math symbols
\usepackage{nicefrac}       % compact symbols for 1/2, etc.
\usepackage{microtype}      % microtypography
\usepackage{lipsum}
\usepackage{fancyhdr}       % header
\usepackage{graphicx}       % graphics
\graphicspath{{media/}}     % organize your images and other figures under media/ folder

\usepackage{hyperref}
\usepackage{url}
\usepackage{booktabs}
\usepackage{amsmath}

\usepackage{amssymb}
\usepackage{xcolor}
\usepackage{graphicx}
\usepackage{tabularx}
\usepackage{wrapfig}
\usepackage{float}
\usepackage{multirow}
\usepackage[numbers]{natbib}
\usepackage{longtable}
\usepackage{titletoc}

\newcommand\mn{{Fisher-IRG}}
\newcommand\cn{{Cov-IRG}}

\title{Fisher-IRG: Fisher-Induced Local Invariant Representation Geometry across \\ Language and Vision Models}

\author{
Abdullah All Tanvir \quad
Xin Zhong
\\ [0.2em]
Department of Computer Science \\
University of Nebraska Omaha, Omaha, NE, USA \\[0.2em]
\texttt{\{atanvir,xzhong\}@unomaha.edu}
}

\begin{document}
\maketitle

\vspace{-2.5em}
\begin{abstract}

Semantic-preserving transformations can induce substantial motion in learned representations, while small changes may strongly affect model predictions, raising a basic question: what local metric best captures semantically consequential variation? We propose Fisher-induced invariant representation geometry (\mn), which measures local representation directions through their predictive sensitivity. Around each representation, we construct semantic-preserving and semantic-changing neighborhoods, aggregate their local Fisher information, and recover invariant directions through a contrastive generalized eigenvalue problem. Controlled displacement analyses first show that comparable Euclidean motion can have substantially different predictive consequences, supporting the need for a predictive geometry. Across language and vision models, \mn\ yields stronger semantic-versus-nuisance predictive selectivity and generally more reproducible subspaces than covariance-based geometry, while recovering systematically distinct local directions. Representation interventions further localize semantic effects to the Fisher-derived subspace, and held-out separation and retrieval show that the recovered geometry generalizes beyond the discovery neighborhoods. These results support \mn\ as a principled framework for characterizing local invariant representation geometry.

\end{abstract}

\vspace{-1.0em}
\section{Introduction}
\label{sec:introduction}
\vspace{-0.75em}

\begin{wrapfigure}{r}{0.69\textwidth}
  \centering
  \vspace{-2.25em}
  \includegraphics[width=1.0\linewidth]{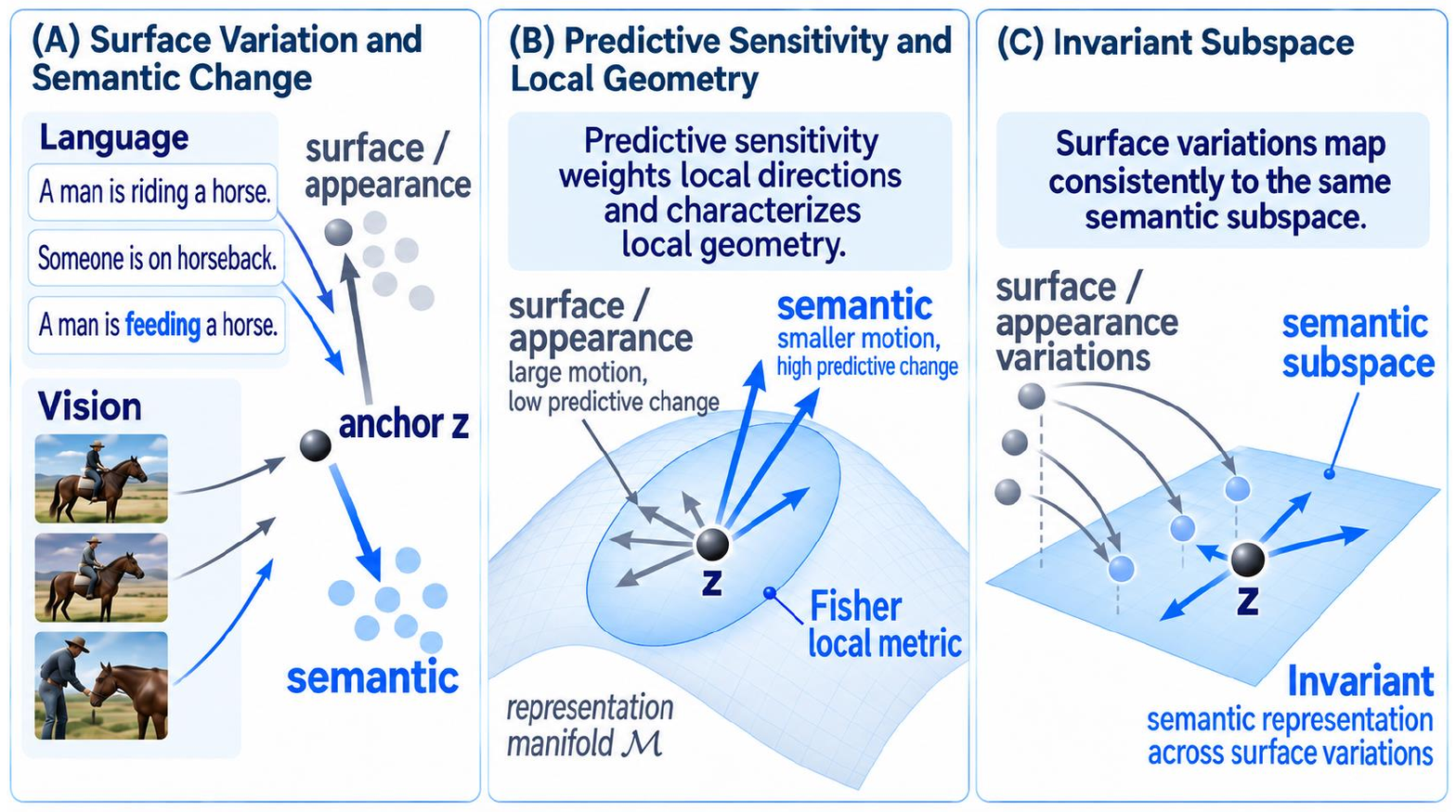}
  \vspace{-9.0em}
    \caption{
    Fisher-induced invariant representation geometry.
    (A) Surface-level variation and semantic change induce different local motions in language and vision representations. (B) Fisher information weights local directions by predictive sensitivity, characterizing the local geometry around an anchor. (C) Contrastive Fisher analysis recovers an invariant subspace in which surface-level variations map consistently while semantically meaningful directions are preserved. 
    }
    \label{fig:PIR}
  \vspace{-1.25em}
\end{wrapfigure}

\vspace{-0.5em}
%%% what we do
Language and vision models must distinguish changes in semantic content from variations in how that content is expressed or observed. Paraphrasing can preserve the meaning of a sentence, just as changes in appearance can preserve the relevant content of an image. Yet stability of meaning does not require every coordinate of a hidden representation to remain unchanged. This raises a geometric question: how should local representation changes be measured when their significance depends on their effects on model behavior? We study this question through Fisher-induced invariant representation geometry, using predictive sensitivity to identify local subspaces that distinguish semantic-changing from semantic-preserving neighborhoods in pretrained language and vision models.

\vspace{-0.5em}
%%% why we need it
The distinction between representation motion and predictive change is central to this problem. Two perturbations with the same Euclidean magnitude can have substantially different effects on a model's output distribution. Likewise, a direction with large displacement variance may reflect frequent nuisance variation, while a direction with small variance may strongly influence prediction. Covariance operators capture the directional structure of sampled motion and can themselves be anisotropic, but they do not directly measure its predictive consequences. A geometric account of semantic invariance therefore requires semantic neighborhoods that specify the variation of interest and a local metric that measures its predictive consequence. In \mn, the neighborhoods define the semantic contrast, while Fisher information supplies the predictive geometry through which that contrast is resolved.

\vspace{-0.5em}
%%% current summary
Existing work provides foundations for these two components. Studies of language representations and visual invariance establish that learned features exhibit structured responses to linguistic and image transformations~\citep{reif2019geometry,lenc2015understanding}, while information-theoretic analyses connect nuisance invariance to representation minimality~\citep{achille2018emergence}. Representation similarity methods characterize relationships between activation spaces~\citep{kornblith2019similarity}, and contrastive spectral methods extract structure enriched in one population relative to another~\citep{abid2018contrastive}. More directly, covariance-based invariant representation analysis separates semantic-changing and semantic-preserving displacement statistics through a local generalized eigenproblem~\citep{dasgupta2026invariantfeatureslanguagemodels}. In parallel, geometric analyses of generative models motivate model-induced latent metrics~\citep{arvanitidis2018latent,arvanitidis2022pulling}, and recent work applies pullback Fisher geometry to activation steering in Transformers~\citep{wang2026fishback}. 
These developments motivate our focus on discovering invariant subspaces by contrasting predictive sensitivity across semantically controlled neighborhoods, using Fisher geometry as the local predictive ruler.

%%% our work
\vspace{-0.5em}
We develop a local geometric framework in which Fisher information measures the predictive significance of representation directions, yielding Fisher-induced local representation geometry (\mn; Fig.~\ref{fig:PIR}). With the model frozen, the Fisher quadratic form captures the leading local change in predictive KL divergence. For each anchor, we aggregate these local forms over semantic-preserving and semantic-changing neighborhoods and solve a regularized generalized eigenproblem to identify directions with greater predictive sensitivity to semantic change than to meaning-preserving variation. This compares local responses across neighborhood states without treating finite anchor-to-variant displacements as infinitesimal tangent vectors. We estimate the operators from first-order predictive score gradients and exploit their low-rank structure for subspace extraction. 

\vspace{-0.5em}
%%% our contributions
Our contributions are threefold.
(1) We formulate local invariant representation analysis through a predictive quadratic form, distinguishing the geometry of sampled representation motion from the geometry of its behavioral consequences.
(2) We develop a contrastive Fisher subspace-discovery procedure with a local KL interpretation, first-order score computation, and low-rank generalized eigendecomposition, applicable to specified predictive readouts in both language and vision models.
(3) We evaluate the resulting geometry across language and vision architectures through predictive selectivity, basis-invariant reproducibility, direct geometric comparison with covariance-based discovery, causal representation interventions, and held-out semantic separation and retrieval. Fisher-IRG yields stronger semantic-versus-nuisance predictive selectivity, generally more reproducible subspaces, and geometrically distinct directions whose intervention selectively affects semantic behavior. These findings support predictive sensitivity as a principled geometric basis for analyzing local semantic invariance in learned representations.

\vspace{-1.5em}
\section{Related Work}
\vspace{-0.5em}
\label{sec:related_work}

\vspace{-0.5em}
\textbf{Semantic Invariance in Language and Vision Representations.}
Analyses of language models reveal structured linguistic information across depth and geometric organization of contextual representations~\citep{tenney2019bert,reif2019geometry}. In vision, \citet{lenc2015understanding} characterize how image transformations affect representations through invariance, equivariance, and equivalence. From an information-theoretic perspective, \citet{achille2018emergence} connect nuisance invariance to the minimality of sufficient representations, providing a theoretical account of how invariant structure can emerge during learning. Complementary methods explicitly shape representations through contrastive objectives: SimCLR learns visual features from augmented views~\citep{chen2020simclr}, while SimCSE learns sentence embeddings using dropout-based views or supervised entailment and contradiction pairs~\citep{gao2021simcse}. These studies motivate the joint requirement of stability under nuisance variation and sensitivity to meaningful differences. Our work examines this requirement locally in pretrained language and vision models, identifying invariant subspaces through the predictive effects of semantic-preserving and semantic-changing variation rather than introducing a representation-training objective.

\vspace{-0.5em}
\textbf{Representation Geometry and Contrastive Subspace Analysis.}
Representation comparison methods, such as SVCCA~\citep{raghu2017svcca}, PWCCA~\citep{morcos2018pwcca}, and CKA~\citep{kornblith2019similarity}, quantify relationships between activation spaces across layers and models. Complementary spectral methods identify structure through contrasts between datasets. In particular, contrastive PCA extracts directions whose variance is enriched in a target dataset relative to a background dataset~\citep{abid2018contrastive}. More directly related, \citet{dasgupta2026invariantfeatureslanguagemodels} formulate local invariant-subspace discovery using covariance operators that contrast semantic-changing and semantic-preserving representation displacements. This formulation provides the local neighborhood perspective and contrastive decomposition on which our analysis builds. We retain this distinction between semantic and nuisance variation while changing the geometric quantity being measured: displacement covariance describes sampled motion in representation space, whereas our Fisher-based construction measures local predictive sensitivity. The resulting subspaces therefore characterize invariance with respect to model behavior, extending the analysis beyond displacement statistics.

\vspace{-0.5em}
\textbf{Fisher Information and Predictive Representation Geometry.}
Fisher information relates local changes in a statistical model to the second-order expansion of KL divergence, providing a foundation for natural-gradient optimization~\citep{martens2020new}. In deep generative models, \citet{arvanitidis2018latent} show that nonlinear generators induce non-Euclidean latent geometry, while \citet{arvanitidis2022pulling} construct latent metrics by pulling back Fisher--Rao geometry from decoder distributions. Related work in language models also emphasizes that geometric operations depend on the choice of inner product: \citet{park2024linear} introduce a causal inner product connecting concept representations, probing, and steering. More recently, FishBack pulls the output Fisher metric back to intermediate Transformer activations and derives steering directions that limit off-target predictive distortion~\citep{wang2026fishback}. These contributions establish the foundations for measuring representation changes through their effects on model distributions. Building on this perspective, our contribution is a contrastive construction of local invariant subspaces from semantic-preserving and semantic-changing neighborhoods. We use Fisher geometry for subspace discovery and analysis across language and vision models, with basis-invariant stability evaluation, rather than for parameter optimization, latent interpolation, or targeted activation steering.

\vspace{-1.5em}
\section{Fisher-Induced Local Invariant Representation Geometry}
\label{sec:method}
\vspace{-1.0em}
% \vspace{-1.0em}
\subsection{Local Representation Geometry and Predictive Sensitivity}
\label{sec:local_geometry}
\vspace{-1.0em}

Let $\Phi_\ell$ denote the representation map at layer $\ell$ of a frozen model, and let $z=\Phi_\ell(x)\in\mathbb{R}^d$ denote the representation of an input $x$. Under the manifold hypothesis~\citep{fefferman2016testing}, we assume that these representations concentrate near a lower-dimensional manifold $\mathcal{M}_\ell\subset\mathbb{R}^d$, with a smooth local structure around the points under analysis. This assumption provides a local model of representation geometry without requiring a global parameterization of the manifold. We suppress the layer index when the context is clear.

\begin{wrapfigure}{r}{0.75\textwidth}
  \centering
  \vspace{-1.5em}
  \includegraphics[width=1.0\linewidth]{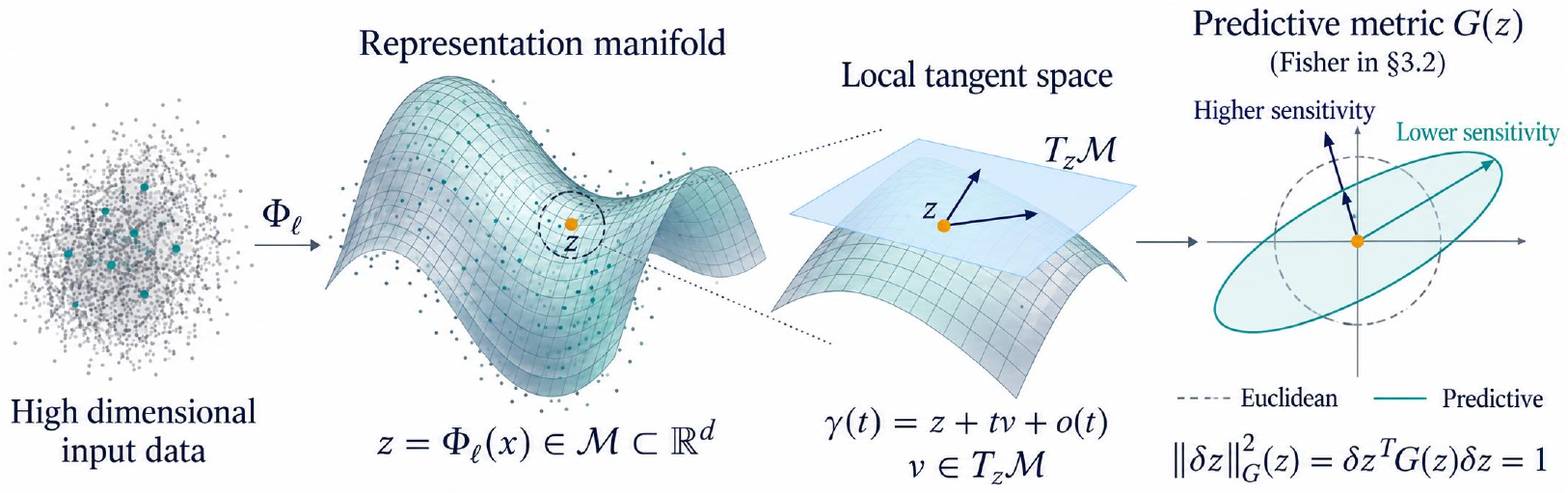}
  \vspace{-18.0em}
    \caption{From representation manifolds to predictive local geometry.
    A model maps high-dimensional inputs onto a representation manifold, whose tangent space captures infinitesimal local motion around an anchor $z$. A predictive metric $G(z)$ then reweights these tangent directions by their effect on model behavior, inducing anisotropic local geometry that differs from the Euclidean metric.}
    \label{fig:theory}
  \vspace{-1.00em}
\end{wrapfigure}

\vspace{-0.5em}
At a local anchor $z\in\mathcal{M}$, the tangent space $T_z\mathcal{M}$ describes the first-order directions of motion along the representation manifold. For a smooth curve $\gamma(t)\in\mathcal{M}$ passing through $\gamma(0)=z$, its initial velocity $v=\dot{\gamma}(0)$ belongs to $T_z\mathcal{M}$ and satisfies
\(
    \gamma(t)=z+tv+o(t).
    \label{eq:local_tangent}
\)
% \begin{equation}
%     \gamma(t)=z+tv+o(t).
%     \label{eq:local_tangent}
% \end{equation}
Local motion can have different semantic roles: some variations preserve the underlying meaning, whereas others modify the content represented by the model. We seek a local representation component that remains stable under the former while retaining sensitivity to the latter. This joint requirement distinguishes semantic invariance from indiscriminate insensitivity. Neither semantic equivalence nor the manifold assumption alone guarantees that equivalent inputs are close under the ambient Euclidean distance.

\vspace{-0.5em}
The tangent space specifies possible local directions, but does not determine their significance for model behavior. To measure this significance, we equip local directions with a symmetric positive-semidefinite operator $G(z)$ and define
\(
    g_z(u,v)=u^\top G(z)v,
    \space 
    \|v\|_{G(z)}^2=v^\top G(z)v,
    \space 
    (u,v\in T_z\mathcal{M}).
    \label{eq:local_predictive_metric}
\)
% \begin{equation}
%     g_z(u,v)=u^\top G(z)v,
%     \qquad
%     \|v\|_{G(z)}^2=v^\top G(z)v,
%     \qquad u,v\in T_z\mathcal{M}.
%     \label{eq:local_predictive_metric}
% \end{equation}
Our desired interpretation is predictive: $g_z(v,v)$ should quantify the leading local change in the model's predictive distribution induced by motion along $v$. Equal Euclidean displacements can then have different magnitudes under $G(z)$, reflecting different effects on the model's output. When $G(z)$ varies smoothly and is positive definite on $T_z\mathcal{M}$, it defines a Riemannian metric. If predictive behavior is insensitive to some nonzero tangent directions, the induced form may instead be degenerate and defines a seminorm.

\vspace{-0.5em}
This distinction has a direct geometric interpretation. In orthonormal tangent coordinates, the Euclidean unit sphere assigns equal length to every unit direction. A positive-definite predictive metric instead defines the constant-sensitivity surface
\(
    \mathcal{E}_z(c)
    =
    \left\{
        v\in T_z\mathcal{M}
        \;:\;
        v^\top G(z)v=c
    \right\},
    (c>0).
    \label{eq:metric_ellipsoid}
\)
% \begin{equation}
%     \mathcal{E}_z(c)
%     =
%     \left\{
%         v\in T_z\mathcal{M}
%         \;:\;
%         v^\top G(z)v=c
%     \right\},
%     \qquad c>0.
%     \label{eq:metric_ellipsoid}
% \end{equation}
This surface is an ellipsoid whose axes depend on the local predictive sensitivity. Longer axes indicate directions along which larger movements are required to produce the same local predictive change; shorter axes indicate greater sensitivity. Its orientation and axis lengths may vary across the representation manifold. Predictive sensitivity is nevertheless model-dependent: whether it distinguishes semantic change from meaning-preserving variation must be established through semantic constraints and empirical evaluation.

\vspace{-0.5em}
\textbf{Euclidean Baseline.} 
Our formulation shares the local manifold perspective of covariance-based invariant representation analysis~\citep{dasgupta2026invariantfeatureslanguagemodels}, which we denote as \cn, while introducing a different object of geometric measurement. \cn\ summarizes the directional variation of sampled representation displacements through local covariance operators, which characterize where representations move but do not directly quantify the predictive consequences of that motion. Here, \mn\ instead introduces a model-induced quadratic form that measures local directions by their effect on predictive behavior. Section~\ref{sec:fisher_discovery} instantiates this form using Fisher information and develops a data-driven procedure for discovering contrastive subspaces.

%%%%
%%%%
%%%%

%%%%
%%%%
%%%% Fisher method

\vspace{-1.25em}
\subsection{Data-Driven Fisher Subspace Discovery}
\label{sec:fisher_discovery}
\vspace{-1.0em}

We instantiate the predictive geometry of Section~\ref{sec:local_geometry} through a data-driven procedure that contrasts local Fisher sensitivity across semantically controlled neighborhoods. The model remains frozen throughout: semantic variants provide the discovery samples, while derivatives of the predictive distribution determine the local geometric operators.

\vspace{-0.5em}
\textbf{Semantically controlled local neighborhoods.}
For an anchor input $x_0$ with representation $z_0=\Phi_\ell(x_0)$, we construct semantic-preserving (SP) and semantic-changing (SC) neighborhoods,
\(
    \mathcal{N}_{\mathrm{sp}}(x_0)
    =
    \{x_i^{\mathrm{sp}}\}_{i=1}^{n_{\mathrm{sp}}}, 
    \space
    \mathcal{N}_{\mathrm{sc}}(x_0)
    =
    \{x_j^{\mathrm{sc}}\}_{j=1}^{n_{\mathrm{sc}}}.
    \label{eq:semantic_neighborhoods}
\)
% \begin{equation}
%     \mathcal{N}_{\mathrm{sp}}(x_0)
%     =
%     \{x_i^{\mathrm{sp}}\}_{i=1}^{n_{\mathrm{sp}}},
%     \qquad
%     \mathcal{N}_{\mathrm{sc}}(x_0)
%     =
%     \{x_j^{\mathrm{sc}}\}_{j=1}^{n_{\mathrm{sc}}}.
%     \label{eq:semantic_neighborhoods}
% \end{equation}
SP variants modify nuisance factors while preserving the semantic content of interest; SC variants introduce controlled changes to that content. Their representations are denoted by $z_i^a=\Phi_\ell(x_i^a)$, where $a\in\{\mathrm{sp},\mathrm{sc}\}$. These sets operationalize semantic variation without assuming that semantic equivalence implies Euclidean proximity. In particular, locality refers to the differential analysis around each sampled representation; we do not approximate every anchor-to-variant displacement as an infinitesimal tangent vector. Discovery neighborhoods are separated from the variants used for subsequent evaluation.

\vspace{-0.5em}
\textbf{Fisher as the local predictive form.}
Let $p_{\ell,x}(y\mid z)$ denote the predictive distribution obtained by varying the analyzed layer representation while holding the remaining context of input $x$ fixed. This notation makes the intervention coordinate explicit: the analyzed vector need not determine the entire model state. We suppress $\ell$ and $x$ when unambiguous. For a discrete output space $\mathcal{Y}$, define the representation score and Fisher information matrix as
% \(
%     g_y(z)
%     =
%     \nabla_z \log p(y\mid z),
%     \space
%     F(z)
%     =
%     \sum_{y\in\mathcal{Y}}
%     p(y\mid z)\,g_y(z)g_y(z)^\top.
%     \label{eq:representation_fisher}
% \)
\vspace{-0.5em}
\begin{equation}
    g_y(z)
    =
    \nabla_z \log p(y\mid z),
    \qquad
    F(z)
    =
    \sum_{y\in\mathcal{Y}}
    p(y\mid z)\,g_y(z)g_y(z)^\top.
    \label{eq:representation_fisher}
\end{equation}

\vspace{-1.0em}
The derivatives are taken with respect to the representation rather than the model parameters. So, $v^\top F(z)v$ measures predictive sensitivity to an intervention along the representation direction $v$.

\vspace{-0.5em}
For a smooth predictive distribution with locally fixed support, normalization gives
$\mathbb{E}_{p}[g_y(z)]=0$ and
$-\mathbb{E}_{p}[\nabla_z^2\log p(y\mid z)]=F(z)$.
A second-order expansion of the predictive KL divergence therefore yields
\(
    D_{\mathrm{KL}}
    \!\left(
        p(\cdot\mid z)
        \,\middle\|\,
        p(\cdot\mid z+\delta z)
    \right)
    =
    \frac{1}{2}\delta z^\top F(z)\delta z
    +
    o(\|\delta z\|^2).
    \label{eq:fisher_local_kl}
\)
% \begin{equation}
%     D_{\mathrm{KL}}
%     \!\left(
%         p(\cdot\mid z)
%         \,\middle\|\,
%         p(\cdot\mid z+\delta z)
%     \right)
%     =
%     \frac{1}{2}\delta z^\top F(z)\delta z
%     +
%     o(\|\delta z\|^2).
%     \label{eq:fisher_local_kl}
% \end{equation}
Thus, Fisher realizes the predictive quadratic form introduced in Section~\ref{sec:local_geometry}: it is the local curvature of distributional change~\citep{martens2020new}. Restricting this form to tangent directions gives the corresponding geometry on the representation manifold. The ambient form also remains well-defined for activation interventions that need not lie exactly on that manifold.

\vspace{-0.5em}
\textbf{Contrastive Fisher operators.}
For each anchor, we aggregate the sample-specific Fisher matrices within the two neighborhoods:
\(
    F_a
    =
    \frac{1}{n_a}
    \sum_{i=1}^{n_a}F(z_i^a),
    \space
    a\in\{\mathrm{sp},\mathrm{sc}\}.
    \label{eq:neighborhood_fisher}
\)
% \begin{equation}
%     F_a
%     =
%     \frac{1}{n_a}
%     \sum_{i=1}^{n_a}F(z_i^a),
%     \qquad
%     a\in\{\mathrm{sp},\mathrm{sc}\}.
%     \label{eq:neighborhood_fisher}
% \end{equation}
All matrices are expressed in the common ambient coordinates of the same model layer. This aggregation compares the response to a shared intervention direction across neighborhood states; it does not require identifying distinct tangent spaces through parallel transport.

\vspace{-0.5em}
To interpret these operators, define the neighborhood-averaged predictive change
\(
    \mathcal{K}_a(t,v)
    =
    \frac{1}{n_a}
    \sum_{i=1}^{n_a}
    D_{\mathrm{KL}}
    \!\left(
        p_i^a(\cdot\mid z_i^a)
        \,\middle\|\,
        p_i^a(\cdot\mid z_i^a+tv)
    \right)
    =
    \frac{t^2}{2}v^\top F_a v+o(t^2),
    \label{eq:neighborhood_kl}
\)
% \begin{equation}
%     \mathcal{K}_a(t,v)
%     =
%     \frac{1}{n_a}
%     \sum_{i=1}^{n_a}
%     D_{\mathrm{KL}}
%     \!\left(
%         p_i^a(\cdot\mid z_i^a)
%         \,\middle\|\,
%         p_i^a(\cdot\mid z_i^a+tv)
%     \right)
%     =
%     \frac{t^2}{2}v^\top F_a v+o(t^2),
%     \label{eq:neighborhood_kl}
% \end{equation}
where $p_i^a=p_{\ell,x_i^a}$ retains the context associated with each sample. We seek directions with greater predictive sensitivity in SC neighborhoods relative to SP neighborhoods through the regularized generalized Rayleigh quotient
\(
    \mathcal{R}_F(v)
    =
    v^\top F_{\mathrm{sc}}v \space
    /
    \space
    v^\top(F_{\mathrm{sp}}+\epsilon I_d)v,
    \space
    v\neq 0,\space 
    \epsilon>0.
    \label{eq:fisher_rayleigh}
\)
% \begin{equation}
%     \mathcal{R}_F(v)
%     =
%     \frac{v^\top F_{\mathrm{sc}}v}
%     {v^\top(F_{\mathrm{sp}}+\epsilon I_d)v},
%     \qquad
%     v\neq 0,\quad \epsilon>0.
%     \label{eq:fisher_rayleigh}
% \end{equation}
The ridge term makes the denominator positive definite even when the estimated Fisher matrices are rank deficient. Maximizing this quotient under
$v^\top(F_{\mathrm{sp}}+\epsilon I_d)v=1$
leads to
\vspace{-0.5em}
\begin{equation}
    F_{\mathrm{sc}}v_r
    =
    \lambda_r(F_{\mathrm{sp}}+\epsilon I_d)v_r,
    \qquad
    \lambda_1\geq\cdots\geq\lambda_d.
    \label{eq:fisher_generalized_eigen}
\end{equation}

\vspace{-0.75em}
We retain the leading $k$ generalized eigenvectors and construct
\(
    V_k=[v_1,\ldots,v_k],
    \space
    U_k=\operatorname{orth}(V_k),
    \space
    P_k=U_kU_k^\top.
    \label{eq:fisher_subspace}
\)
% \begin{equation}
%     V_k=[v_1,\ldots,v_k],
%     \qquad
%     U_k=\operatorname{orth}(V_k),
%     \qquad
%     P_k=U_kU_k^\top.
%     \label{eq:fisher_subspace}
% \end{equation}
Here, $\operatorname{orth}$ returns a Euclidean-orthonormal basis of the same span, so $P_k$ defines a basis-independent projection. The discovered subspace identifies directions of relative predictive selectivity across the two neighborhoods. This criterion does not, by itself, establish stability of projected representations under finite SP transformations; we assess that semantic interpretation through held-out representation and intervention experiments.

\vspace{-0.5em}
\textbf{First-order computation and output truncation.}
Although Fisher describes second-order predictive change, its score representation in Eq.~\ref{eq:representation_fisher} requires only first-order derivatives. We compute these gradients by differentiating the frozen predictive readout with respect to its representation input, without constructing a representation Hessian.

\vspace{-0.5em}
For large output spaces, we retain the $q$ highest-probability outcomes at each sample. Let $\mathcal{T}_q(z)\subseteq\mathcal{Y}$ denote this set and let
$m_q(z)=\sum_{y\in\mathcal{T}_q(z)}p(y\mid z)$
be its retained probability mass. Following this selection, we form
\(
    \widetilde{p}_y(z)
    =
    p(y\mid z) \space / \space  m_q(z),
    \space
    \widehat{F}_q(z)
    =
    \sum_{y\in\mathcal{T}_q(z)}
    \widetilde{p}_y(z)\,
    g_y(z)g_y(z)^\top.
    \label{eq:topq_fisher}
\)
% \begin{equation}
%     \widetilde{p}_y(z)
%     =
%     \frac{p(y\mid z)}{m_q(z)},
%     \qquad
%     \widehat{F}_q(z)
%     =
%     \sum_{y\in\mathcal{T}_q(z)}
%     \widetilde{p}_y(z)\,
%     g_y(z)g_y(z)^\top.
%     \label{eq:topq_fisher}
% \end{equation}
The scores $g_y(z)$ are computed from the original predictive distribution; output selection and weight renormalization are used only to aggregate their outer products. Accordingly, $\widehat{F}_q$ is a truncated approximation to the full Fisher, rather than the exact Fisher of a softmax restricted to the selected outcomes. It recovers $F$ when all outcomes are retained. Approximation quality depends on both omitted probability mass and the associated score magnitudes, motivating sensitivity analysis with respect to $q$.

\vspace{-0.5em}
\textbf{Low-rank subspace computation.}
Define the sample factor
\(
    B_i^a
    =
    \left[
        \sqrt{\widetilde{p}_y(z_i^a)}\,
        g_y(z_i^a)
    \right]_{y\in\mathcal{T}_q(z_i^a)},
    \space
    A_a
    =
    \frac{1}{\sqrt{n_a}}
    [B_1^a,\ldots,B_{n_a}^a].
    \label{eq:fisher_factors}
\)
% \begin{equation}
%     B_i^a
%     =
%     \left[
%         \sqrt{\widetilde{p}_y(z_i^a)}\,
%         g_y(z_i^a)
%     \right]_{y\in\mathcal{T}_q(z_i^a)},
%     \qquad
%     A_a
%     =
%     \frac{1}{\sqrt{n_a}}
%     [B_1^a,\ldots,B_{n_a}^a].
%     \label{eq:fisher_factors}
% \end{equation}
The approximate neighborhood operator then satisfies
$\widehat{F}_a=A_aA_a^\top$,
with rank at most $\min(d,n_aq)$.
Its action on a vector is evaluated as
$\widehat{F}_av=A_a(A_a^\top v)$,
avoiding explicit storage of a dense $d\times d$ matrix.

\vspace{-0.5em}
For an exact reduction of the approximate eigenproblem, let $Q$ be an orthonormal basis of
$\operatorname{span}([A_{\mathrm{sp}},A_{\mathrm{sc}}])$.
Every generalized eigenvector with a nonzero eigenvalue lies in this span and can be written as $v=Qw$. We therefore solve
\begin{equation}
    (Q^\top\widehat{F}_{\mathrm{sc}}Q)w
    =
    \lambda
    \left(
        Q^\top\widehat{F}_{\mathrm{sp}}Q+\epsilon I_r
    \right)w,
    \qquad r=\operatorname{rank}(Q),
    \label{eq:reduced_fisher_eigen}
\end{equation}
and recover the ambient directions through $v=Qw$. When the factor span is large, the same factorization supports iterative generalized eigensolvers through matrix--vector products. Thus, computation adapts to the effective rank of the sampled predictive gradients.

\vspace{-0.5em}
\textbf{Language and vision readouts.}
The construction requires a differentiable predictive distribution and a specified representation intervention. For language models, the output space is the vocabulary, and the readout produces the next-token distribution at the final valid prompt position. A token-level intervention varies the selected hidden state while keeping the remaining states fixed. For a mean-pooled representation, the intervention must additionally specify how a displacement of the pooled vector is lifted to the token states; pooling alone does not define the downstream predictive map.

\vspace{-0.5em}
For vision models, the predictive distribution is obtained by applying a softmax to a fixed set of class or candidate scores,
\(
    p_{\ell,x}(y\mid z)
    =
    \exp(s_{\ell,x,y}(z)/\tau) 
    \space / \space
    \sum_{c\in\mathcal{Y}}
    \exp(s_{\ell,x,c}(z)/\tau),
    \label{eq:vision_predictive_readout}
\)
where $s_{\ell,x,y}$ includes the remaining computation and the specified readout, and $\tau>0$ is its temperature. The head, candidate set, temperature, and intervention rule are held fixed within each analysis and specified in the experimental setup. For encoders without a native categorical output, the chosen readout is part of the definition of the measured geometry. Across both modalities, the same predictive map and intervention convention are used for Fisher estimation and subsequent predictive evaluation.

%%%%
%%%%
%%%%

\vspace{-1.50em}
\section{Experiments}
\label{sec:experiments}
\vspace{-1.00em}
\subsection{Datasets and Model Coverage}
\label{sec:exp_setup}
\vspace{-0.5em}

\vspace{-0.5em}
\textbf{Datasets.} 
We construct controlled SP and SC neighborhoods for both language and vision, enabling the same geometric hypothesis to be evaluated across modalities. The text dataset contains 1,800 anchor prompts and 135,000 sentences in total. Each anchor is paired with 10 SP variants that preserve meaning while varying lexical choice, syntax, and phrasing, together with 64 SC variants that modify the underlying semantic content. The SC variants are divided into easy and hard conditions to cover changes of different character and difficulty. For example, from the anchor ``A doctor examined the patient in a quiet clinic before prescribing antibiotics,'' an SP variant is ``Before prescribing antibiotics, the physician examined the patient in a quiet clinic.'' An easy SC variant changes a small number of explicit semantic elements, e.g., ``A nurse examined the patient in a quiet clinic before prescribing pain medication,'' whereas a hard SC variant introduces broader semantic changes, e.g., ``A pharmacist advised a customer in a crowded store before recommending an allergy treatment.''

\begin{wrapfigure}{r}{0.5\textwidth}
  \centering
  \vspace{-1.5em}
  \includegraphics[width=1.0\linewidth]{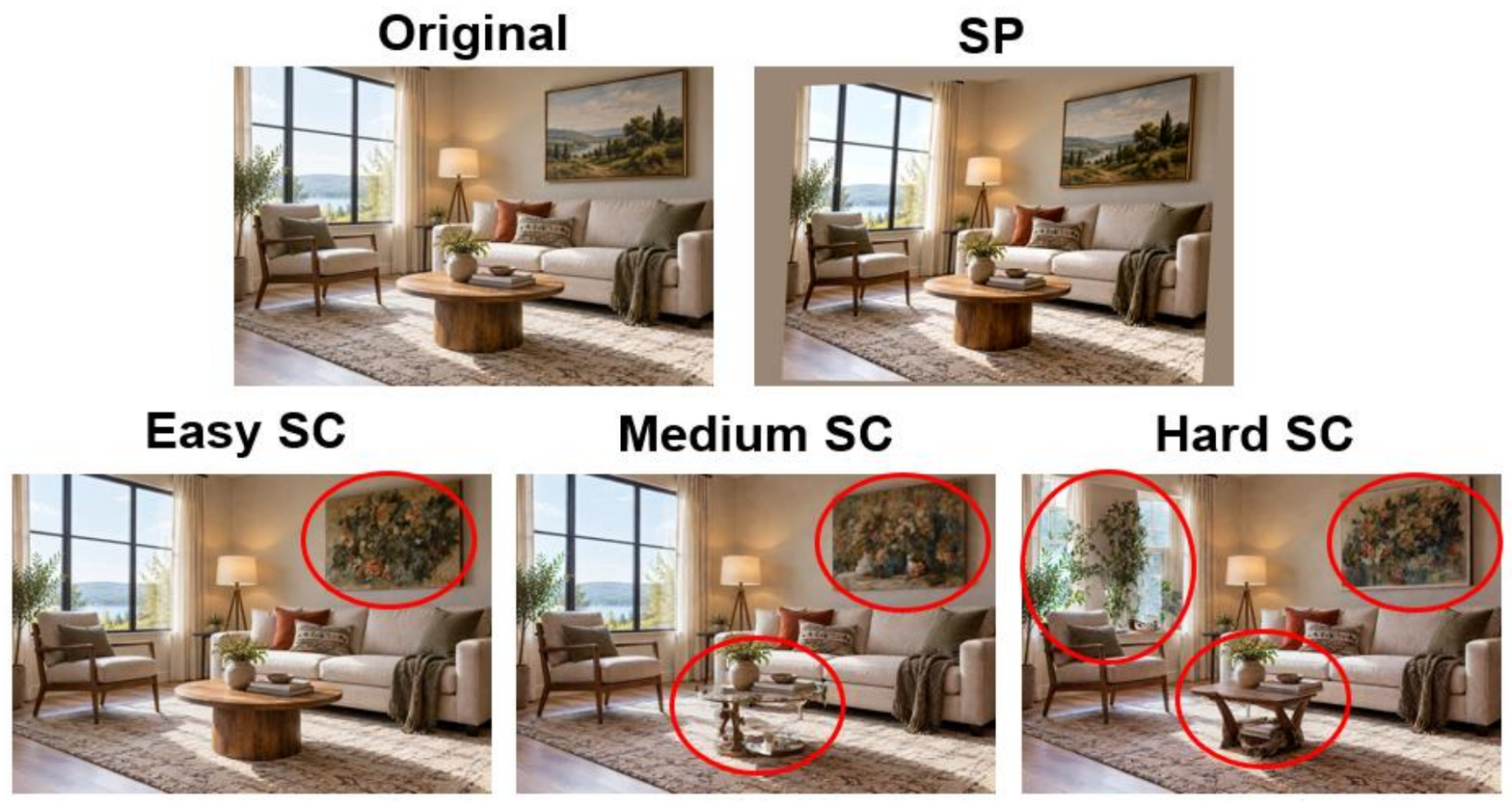}
  \vspace{-7.0em}
    \caption{Example visual neighborhood construction.
    SP variants preserve the source semantics, whereas easy, medium, and hard SC variants introduce progressively larger semantic modifications. Red circles manually highlight the changes.}
    \label{fig:main_img_eg}
  \vspace{-1.00em}
\end{wrapfigure}

\vspace{-0.5em}
For vision, we construct neighborhoods around 1,000 source images selected from Visual Genome. Each neighborhood contains one anchor, 10 SP variants, and 64 SC variants spanning easy, medium, and hard semantic-change conditions, yielding 75,000 image--description instances in total. SP variants preserve the principal objects, attributes, relations, and semantic content of the source, whereas SC variants introduce controlled semantic modifications of varying difficulty. Fig.~\ref{fig:main_img_eg} shows one example neighborhood; semantic changes are manually highlighted with red circles for readability. 

\vspace{-0.5em}
For both modalities, the available variants are partitioned into disjoint discovery and evaluation subsets: only the discovery subset is used to estimate the Fisher and covariance subspaces, while held-out variants are used for subsequent evaluation. This separation prevents the reported semantic selectivity and stability from being measured on the same perturbations used to discover the geometry. Further construction and split details are provided in Appendix~\ref{app:data}.

\vspace{-0.5em}
\textbf{Models.}
We evaluate the framework on seven language models spanning distinct model families, parameter scales, and training designs: Mistral-Instruct-v0.3, LLaMA-3-Instruct, Gemma-IT, Qwen2.5-Instruct, DeepSeek-MoE-Chat, Phi-4-Mini, and Falcon3-Instruct. The collection ranges from 4B to 16B parameters and includes dense and mixture-of-experts architectures as well as models obtained through different instruction-tuning and alignment pipelines. For vision, we evaluate ViT-B/16, DINOv2 ViT-B/14, and CLIP ViT-B/16. Although all three use transformer-based visual representations, they are learned under substantially different objectives: supervised recognition, self-supervised representation learning, and image--text contrastive learning, respectively. This model suite allows us to test whether Fisher-induced invariant geometry persists across architectures, training paradigms, and modalities rather than arising from a particular model family. Complete model and implementation details are provided in Appendix~\ref{app:implementation}.

%%%
%%% Displacement-Consequence Separability
%%% Decouple

\vspace{-1.25em}
\subsection{Displacement-Consequence Separability}
\label{sec:euclidean_predictive}
\vspace{-0.5em}

\begin{wraptable}{r}{0.37\textwidth}
\vspace{-1.75em}
\centering
\small
\caption{Predictive-consequence mismatch among distance-matched perturbations. Rate is the fraction of matched cases with $\rho_{\mathrm{KL}}\ge 2$; Layers count layers containing such cases.}
\label{tab:euclidean_predictive_summary}
\setlength{\tabcolsep}{3.6pt}
\begin{tabular}{lccc}
\toprule
Model & Rate & Layers & $\rho_{\mathrm{KL}}$ \\
\midrule
\multicolumn{2}{l}{\textit{Language}} \\
Mistral  & 24.1\% & 32/32 & 2.03--16.99$\times$ \\
Qwen  & 12.6\% & 27/28 & 2.84--7.57$\times$ \\
LLaMA  & 5.8\%  & 22/32 & 2.03--4.27$\times$ \\
Gemma    & 17.1\% & 28/28 & 2.40--187.50$\times$ \\
\midrule
\multicolumn{2}{l}{\textit{Vision}} \\
ViT-B/16 & 11.1\% & 12/12 & 2.43--3.59$\times$ \\
\bottomrule
\end{tabular}
\vspace{-1.0em}
\end{wraptable}

\vspace{-0.5em}
For a held-out perturbation with representation change
\(
\Delta z=z'-z,
\)
the Euclidean quantity $\|\Delta z\|_2$ measures how far the representation moves in latent space, while
\(
D_{\mathrm{KL}}\!\left(p(z)\,\|\,p(z+\Delta z)\right)
\)
measures the resulting change in the model's predictive distribution. If Euclidean magnitude fully determined local behavioral significance, perturbations with similar $\|\Delta z\|_2$ should induce similar predictive changes. We therefore match held-out SP and SC perturbations within each semantic group and layer by Euclidean displacement. A pair is distance-matched when its relative displacement difference is at most $10\%$. For each matched pair, we compute the predictive-consequence ratio
\(
\rho_{\mathrm{KL}}
=
D_{\mathrm{KL}}^{\mathrm{SC}}/
D_{\mathrm{KL}}^{\mathrm{SP}},
\)
and count a mismatch when $\rho_{\mathrm{KL}}\geq2$. Thus, the analysis isolates cases in which SP and SC perturbations move nearly the same Euclidean distance but induce substantially different predictive effects. Complete layer-wise results are reported in Appendix~\ref{app:euclidean_predictive}.

\vspace{-0.5em}
\textbf{Euclidean displacement magnitude does not uniquely determine predictive consequence, and the mismatch appears across both language and vision representations.}
Table~\ref{tab:euclidean_predictive_summary} shows that distance-matched SP/SC perturbations can induce sharply different predictive changes in every evaluated model. The phenomenon is local rather than universal, but occurs across substantial portions of model depth: all evaluated layers for Mistral, Gemma, and ViT-B/16, 27 of 28 layers for Qwen2.5, and 22 of 32 for LLaMA-3. The reported KL ratios summarize the layer-wise median values among qualifying cases; for example, a value of $3\times$ means that the SC perturbation changes the predictive distribution about three times as much as its Euclidean-matched SP counterpart. In vision, these medians range from $2.43\times$ to $3.59\times$ across layers. These results do not imply that Euclidean displacement and predictive change are globally unrelated; rather, they establish that Euclidean magnitude alone is insufficient to determine local behavioral significance. This directly motivates Fisher-IRG, which measures representation directions through their predictive consequences rather than displacement magnitude alone.

%%%
%%%
%%%

%%%
%%%
%%% Predictive Selectivity and Subspace Stability

\vspace{-1.25em}
\subsection{Predictive Selectivity and Subspace Stability}
\label{sec:selectivity_stability}
\vspace{-0.9em}

\begin{wraptable}{r}{0.20\textwidth}
\vspace{-3.00em}
\centering
\small
\caption{Predictive-selectivity advantage of Fisher-IRG over Cov-IRG. Mean $\Delta S$ is averaged over five normalized depths.}
\label{tab:selectivity_summary}
\setlength{\tabcolsep}{4pt}
\begin{tabular}{lc}
\toprule
Model & Mean $\Delta S \uparrow$ \\
\midrule
\multicolumn{2}{l}{\textit{Language}} \\
LLaMA      & 0.218 \\
Mistral    & 0.230 \\
Gemma      & 0.170 \\
Qwen       & 0.368 \\
DeepSeek   & 0.248 \\
Phi        & 0.222 \\
Falcon     & 0.234 \\
\midrule
\multicolumn{2}{l}{\textit{Vision}} \\
ViT        & 0.628 \\
DINOv2     & 0.516 \\
CLIP       & 0.476 \\
\bottomrule
\end{tabular}
\vspace{-2.50em}
\end{wraptable}

We evaluate whether Fisher-IRG improves both the semantic selectivity and the reproducibility of the discovered invariant geometry relative to Cov-IRG. These properties are complementary: the recovered subspace should respond more strongly to semantic-changing than semantic-preserving variation, and it should remain stable when estimated from independent samples of the same local neighborhood.

\vspace{-0.5em}
\textbf{Predictive selectivity.}
For each semantic group $g$, layer $\ell$, and geometry $m\in\{\mathrm{Fisher},\mathrm{Cov}\}$, let $U^m_{g,\ell}$ denote the discovered subspace. 
Given an anchor $z^{(a)}$ and held-out donor $z^{(d)}$, we intervene by replacing only the component lying in the method-$m$ subspace:
\(
z_{\mathrm{int}}^{m}
=
U^mU^{m\top}z^{(d)}
+
(I-U^mU^{m\top})z^{(a)}.
\)
Thus, the donor contributes only its projection onto the discovered subspace, while the anchor's complementary component is held fixed; $z_{\mathrm{int}}^{m}$ is then propagated through the remaining frozen model. 
Let $D^{m,\mathrm{SC}}_{g,\ell}$ and $D^{m,\mathrm{SP}}_{g,\ell}$ denote the KL divergence between the original anchor predictive distribution and the predictive distribution produced after replacing the method-$m$ subspace component with that from an SC or SP donor, respectively. 
We define
\(
S^{m}_{g,\ell}
=
\log[
(D^{m,\mathrm{SC}}_{g,\ell}+\delta)
/
(D^{m,\mathrm{SP}}_{g,\ell}+\delta)
],
\)
so larger values indicate stronger predictive sensitivity to semantic than meaning-preserving variation. We compare the two geometries by
\(
\Delta S_{g,\ell}
=
S^{\mathrm{Fisher}}_{g,\ell}
-
S^{\mathrm{Cov}}_{g,\ell},
\)
where $\Delta S>0$ favors Fisher-IRG. We evaluate five approximately evenly spaced normalized depths $\ell/L$ with $k=16$, using mean-pooled language and CLS-token vision representations. Full depth-wise results are reported in Appendix~\ref{app:localization}.

\vspace{-0.5em}
\textbf{Fisher-IRG consistently improves predictive selectivity across modalities and depth.}
Table~\ref{tab:selectivity_summary} shows positive $\Delta S$ at every reported depth for all seven language models and all three vision models. Across language models, the mean gain ranges from $0.170$ for Gemma to $0.368$ for Qwen2.5, while the vision models show larger positive margins, from $0.476$ for CLIP to $0.628$ for ViT-B/16. The location of the strongest advantage is architecture dependent: some models peak in early or intermediate layers, while others peak later. Thus, Fisher-IRG does not imply a universal invariant layer; rather, it provides a consistently more selective local geometry whose strongest expression depends on the model.

\vspace{-0.5em}
\textbf{Basis-invariant subspace stability.}
Selectivity alone does not establish that the recovered geometry is reproducible. We therefore ask whether similar invariant subspaces are obtained when the local operators are estimated from independent samples of the discovery neighborhoods. 

\begin{wraptable}{r}{0.24\textwidth}
\vspace{-2.75em}
\centering
\small
\caption{Grassmann reproducibility of Fisher-IRG and Cov-IRG. Distances are averaged over three layers and $k\in\{8,16,32\}$.}
\label{tab:grassmann_summary}
\setlength{\tabcolsep}{3.8pt}
\begin{tabular}{lcc}
\toprule
Model & Fisher $\downarrow$ & Cov $\downarrow$ \\
\midrule
\multicolumn{3}{l}{\textit{Language}} \\
Mistral  & 2.459 & 2.535 \\
LLaMA    & 2.553 & 2.644 \\
Gemma    & 2.732 & 2.781 \\
Qwen     & 2.672 & 2.811 \\
DeepSeek & 2.331 & 2.456 \\
Phi      & 2.337 & 2.427 \\
Falcon   & 2.963 & 3.056 \\
\midrule
\multicolumn{3}{l}{\textit{Vision}} \\
ViT      & 3.283 & 3.370 \\
DINOv2   & 3.311 & 3.455 \\
CLIP     & 3.304 & 3.446 \\
\bottomrule
\end{tabular}
\vspace{-2.0em}
\end{wraptable}

\vspace{-0.5em}
Since the recovered object is a subspace rather than a particular eigenvector basis, we evaluate stability through the Grassmann geometry of $k$-dimensional subspaces.
For two independently estimated subspaces with Euclidean-orthonormal bases $U,\widetilde U\in\mathbb{R}^{d\times k}$, the principal angles satisfy
\(
\theta_i=\cos^{-1}\sigma_i(U^\top\widetilde U),
\ i=1,\ldots,k,
\)
and their Grassmann geodesic distance is~\citep{edelman1998geometry}
\(
d_{\mathrm{Gr}}(U,\widetilde U)
=
\left(\sum_{i=1}^{k}\theta_i^2\right)^{1/2}.
\)
Because this quantity depends only on the two spans, it is invariant to basis sign, ordering, and orthogonal rotation. Smaller $d_{\mathrm{Gr}}$ therefore indicates greater reproducibility of the recovered subspace. 
For each model, layer, and $k\in\{8,16,32\}$, we independently estimate Fisher-IRG and Cov-IRG from different sampled subsets of the discovery neighborhoods and compare them under matched conditions. Results are averaged across semantic groups and independent run pairs at three representative layers; complete layer--dimension values are reported in Appendix~\ref{app:grassmann_stability}. Because $d_{\mathrm{Gr}}$ depends on subspace dimension, Fisher-IRG and Cov-IRG are compared at matched $k$.

\vspace{-0.5em}
\textbf{Fisher-IRG generally recovers a more reproducible subspace than Cov-IRG across models.} The result is reflected by the lower average Grassmann distances in Table~\ref{tab:grassmann_summary}. We do not expect predictive weighting to dominate every finite-sample configuration: reproducibility also depends on the sampled local neighborhood, the effective generalized eigengap, and the representation geometry at a particular layer and dimension. The relevant claim is therefore a systematic cross-model tendency toward lower Grassmann distance rather than uniform dominance at every setting. Complete layer--dimension results are reported in Appendix~\ref{app:grassmann_stability}. 

%%%
%%%
%%%

%%%
%%%
%%% Fisher Cov different

\vspace{-1.25em}
\subsection{Fisher and Covariance Induce Distinct Local Geometries}
\label{sec:distinct_geometry}
\vspace{-0.5em}

\vspace{-0.5em}
The preceding results show that Fisher-IRG provides stronger predictive selectivity and generally greater subspace reproducibility than Cov-IRG. These gains could still arise if Fisher weighting simply produced a better estimate of approximately the same covariance-derived subspace. We therefore test whether the two methods recover geometrically distinct local structures.

\begin{wraptable}{r}{0.325\textwidth}
\vspace{-2.0em}
\centering
\small
\caption{Geometric separation between Fisher- and Cov-IRG. $d(F,C)$ is the cross-method projection distance; $\Delta_{\mathrm{geo}}$ is its excess over within-method variation.}
\label{tab:distinct_geometry}
\setlength{\tabcolsep}{2.2pt}
\begin{tabular}{@{}lccc@{}}
\toprule
Model &
{\scriptsize $d(F,C)\uparrow$} &
{\scriptsize Prin.\ Cos.$\downarrow$} &
{\scriptsize $\Delta_{\mathrm{geo}}\uparrow$} \\
\midrule
\multicolumn{4}{@{}l}{\textit{Language}} \\
Mistral   & .998 & .064 & .559 \\
LLaMA     & .997 & .070 & .484 \\
Gemma     & .998 & .069 & .549 \\
Qwen      & .997 & .056 & .559 \\
DeepSeek  & .998 & .071 & .507 \\
Phi       & .993 & .069 & .581 \\
Falcon    & .993 & .071 & .535 \\
\midrule
\multicolumn{4}{@{}l}{\textit{Vision}} \\
ViT       & .997 & .080 & .479 \\
DINOv2    & .995 & .068 & .529 \\
CLIP      & .992 & .065 & .465 \\
\bottomrule
\end{tabular}
\vspace{-1.25em}
\end{wraptable}

\vspace{-0.5em}
For each semantic group, we independently estimate two Fisher-IRG bases, $U_F,U'_F$, and two Cov-IRG bases, $U_C,U'_C$, from independent discovery splits. For orthonormal bases $U$ and $V$, we measure normalized projection distance
\(
d_{\mathrm{proj}}(U,V)
=
\|UU^\top-VV^\top\|_{F}/\sqrt{2k},
\)
where $0$ denotes identical subspaces and $1$ denotes orthogonal subspaces. We compare the cross-method distance
\(
d_{\mathrm{proj}}(U_F,U_C)
\)
with the within-method distances
\(
d_{\mathrm{proj}}(U_F,U'_F)
\)
and
\(
d_{\mathrm{proj}}(U_C,U'_C),
\)
and define
\(
\Delta_{\mathrm{geo}}
=
d_{\mathrm{proj}}(U_F,U_C)
-
\max\!\left\{
d_{\mathrm{proj}}(U_F,U'_F),
d_{\mathrm{proj}}(U_C,U'_C)
\right\}.
\)
Thus, $\Delta_{\mathrm{geo}}>0$ indicates that changing the geometric operator alters the recovered subspace more than independent re-estimation of either method. 
We also report the mean principal cosine, obtained from the singular values of $U_F^\top U_C$, as a complementary measure of subspace overlap: values near zero indicate little directional overlap. 
Results use $k=16$ at a representative layer; complete analyses are provided in Appendix~\ref{app:distinct_geometry}.

\vspace{-0.5em}
\textbf{Fisher-IRG and Cov-IRG recover systematically different local subspaces.}
As shown in Table~\ref{tab:distinct_geometry}, the cross-method distance exceeds both within-method distances for every evaluated semantic group. For language models, $d(F,C)$ ranges from $0.993$ to $0.998$, with mean principal cosine between $0.056$ and $0.071$. Vision shows the same pattern, with $d(F,C)$ between $0.992$ and $0.997$ and principal cosine between $0.065$ and $0.080$. In every model, $\Delta_{\mathrm{geo}}>0$. 
These results show that Fisher-IRG and Cov-IRG recover systematically different local geometries, rather than alternative estimates of the same subspace.

%%%
%%%
%%%

%%%
%%%
%%% Causal

\vspace{-1.25em}
\subsection{Causal Validation by Representation Intervention}
\label{sec:causal_intervention}
\vspace{-0.5em}

\vspace{-0.5em}
We ask whether the discovered subspace functionally carries semantic information. We replace components of an anchor representation with those from SP or SC donors and measure how much the model output changes.

\vspace{-0.5em}
Let the intervention-layer representation be decomposed as
\(
z=z_{\mathrm{inv}}+z_{\mathrm{nuis}},
\space
z_{\mathrm{inv}}=UU^\top z,
\space
z_{\mathrm{nuis}}=(I-UU^\top)z,
\)
where $U$ spans the Fisher-IRG subspace. Given anchor $z^{(a)}$ and donor $z^{(d)}$, we replace either the Fisher-IRG component or its complement with the corresponding donor component, keep the remaining anchor component fixed, and propagate the intervened representation through the frozen model. For donor type $x\in\{\mathrm{SP},\mathrm{SC}\}$ and replaced component $c\in\{\mathrm{inv},\mathrm{nuis}\}$, $D_{x,c}$ denotes the KL divergence between the original anchor output distribution and the output distribution after intervention. 

\begin{wraptable}{r}{0.40\textwidth}
\vspace{-1.75em}
\centering
\small
\caption{Causal localization under Fisher-IRG intervention. Mean divergences summarize intervention strength; counts and ranges summarize the two causal gaps.}
\label{tab:causal_summary}
\setlength{\tabcolsep}{3.0pt}

\begin{tabular}{lccc}
\toprule
Setting &
$\overline{D}_{\mathrm{SP,inv}}$ &
$\overline{D}_{\mathrm{SC,nuis}}$ &
$\overline{D}_{\mathrm{SC,inv}}$ \\
\midrule
Language & $.003$ & $2.040$ & $3.790$ \\
Vision   & $.002$ & $.187$  & $.571$ \\
\bottomrule
\end{tabular}

\vspace{0.35em}

\begin{tabular}{lcc}
\toprule
Setting & $\Delta_{\mathrm{causal}}>0$ & $\Delta_{\mathrm{local}}>0$ \\
\midrule
Language & 21/21 & 21/21 \\
Vision   & 9/9   & 9/9 \\
\bottomrule
\end{tabular}

\vspace{0.25em}
{\scriptsize
Ranges:
$\Delta_{\mathrm{causal}}$ = .084--9.802 (Lang.), .104--1.005 (Vis.);
$\Delta_{\mathrm{local}}$ = .010--5.076 (Lang.), .070--.794 (Vis.).}
\vspace{-0.8em}
\end{wraptable}

\vspace{-0.5em}
We then define
\(
\Delta_{\mathrm{causal}}
=
D_{\mathrm{SC},\mathrm{inv}}
-
D_{\mathrm{SP},\mathrm{inv}},
\space
\Delta_{\mathrm{local}}
=
D_{\mathrm{SC},\mathrm{inv}}
-
D_{\mathrm{SC},\mathrm{nuis}}.
\)
The first asks whether semantic replacement through Fisher-IRG changes the output more than meaning-preserving replacement; the second asks whether the same semantic donor has greater effect through Fisher-IRG than through its complement. Positive values for both constitute the expected causal-localization pattern.
For language, we use three fixed semantic groups shared across all seven models. For vision, we evaluate ViT-B/16 on independent Visual Genome groups under easy, medium, and hard semantic changes. All experiments use $k=16$ at layer 10; complete group-level results are reported in Appendix~\ref{app:causal_intervention}.

\vspace{-0.5em}
\textbf{Semantic effects are causally concentrated in the Fisher-IRG subspace.}
Table~\ref{tab:causal_summary} shows that both causal-localization criteria are positive for all 21 language model--group and all nine vision group--difficulty combinations. Within Fisher-IRG, SP replacement produces only a small predictive change, whereas SC replacement is substantially more consequential. For the same SC donor, intervention through Fisher-IRG also produces a larger effect than through the complementary component, showing that the effect is not explained solely by generic sensitivity to representation replacement in these settings. 
The pattern holds across all evaluated language architectures and across easy, medium, and hard visual semantic changes. 
These interventions therefore provide functional evidence that predictive semantic sensitivity is concentrated in the Fisher-induced subspace. 

%%%
%%%
%%%

%%%
%%%
%%% held out + retrival

\vspace{-1.25em}
\subsection{Functional Validation of Fisher-IRG}
\label{sec:functional_validation}
\vspace{-0.5em}

\vspace{-0.5em}
We test whether the Fisher-induced geometry generalizes beyond the samples used to construct it. We consider two complementary checks: held-out semantic separation, which evaluates unseen SP/SC perturbations, and semantic retrieval, which tests whether the projected representation supports direct matching against semantic distractors.

\begin{wraptable}{r}{0.43\textwidth}
\vspace{-2.0em}
\centering
\small
\caption{Held-out semantic separation. Wins counts settings in which Fisher-IRG achieves the largest $\Delta_{\mathrm{sep}}$ among Fisher-IRG, Cov-IRG, and the original representation.}
\label{tab:heldout_summary}
\setlength{\tabcolsep}{4.0pt}
\begin{tabular}{lcc}
\toprule
Setting & Fisher Wins & Fisher $\Delta_{\mathrm{sep}}$ Range \\
\midrule
Language & 21/21 & $.000434$--$.021707$ \\
Vision   & 9/9   & $.026423$--$.724569$ \\
\bottomrule
\end{tabular}
\vspace{-1.5em}
\end{wraptable}

\vspace{-0.5em}
\textbf{Fisher-IRG consistently improves held-out semantic separation.}
For each semantic group, SP and SC variants are divided into disjoint discovery and evaluation subsets. The former estimate the subspace, while all reported distances use held-out variants. Let $U$ denote the basis of the discovered Fisher-IRG or comparison subspace, and project each representation as
\(
z_{\mathrm{proj}}=UU^\top z.
\)
We then compute cosine distance in the projected space between the anchor and each held-out SP or SC variant, yielding $d_{\mathrm{SP}}$ and $d_{\mathrm{SC}}$, respectively. We define
\(
\Delta_{\mathrm{sep}}
=
d_{\mathrm{SC}}-d_{\mathrm{SP}}.
\)
A larger positive margin indicates that the projected geometry keeps meaning-preserving variants closer to the anchor while separating semantic-changing variants more strongly.

\vspace{-0.5em}
Across seven language models and three representative layers, Fisher-IRG achieves the largest $\Delta_{\mathrm{sep}}$ in all 21 model--layer comparisons; in vision, it does so in all nine ViT-B/16 layer--difficulty settings (Table~\ref{tab:heldout_summary}). The source of the gain differs across modalities: language improvements are often driven primarily by increased SC separation, whereas later vision layers exhibit both reduced SP variation and increased SC distance. Notably, at the earliest evaluated vision layer, the original and Cov-IRG representations yield negative margins for all three SC difficulty levels, while Fisher-IRG already yields positive separation. 
Complete results are reported in Appendix~\ref{app:functional_validation}.

\begin{wraptable}{r}{0.47\textwidth}
\vspace{-2.0em}
\centering
\small
\caption{Held-out semantic retrieval averaged across seven language models.}
\label{tab:functional_retrieval}
\setlength{\tabcolsep}{4.0pt}
\begin{tabular}{lccc}
\toprule
Representation
& Pairwise $\uparrow$
& MRR $\uparrow$
& Hit@1 $\uparrow$ \\
\midrule
Original   & .949 & .921 & .873 \\
Random     & .916 & .876 & .803 \\
PCA        & .935 & .896 & .829 \\
cPCA       & .945 & .905 & .841 \\
Cov-IRG    & .914 & .872 & .791 \\
Fisher-IRG & \textbf{.959} & \textbf{.931} & \textbf{.885} \\
\bottomrule
\end{tabular}
\vspace{-1.25em}
\end{wraptable}

\vspace{-0.5em}
\textbf{The held-out geometry also supports stronger semantic retrieval.}
For each of the 1,800 language semantic groups, the anchor serves as a query, one held-out SP variant is the relevant candidate, and five held-out SC variants are distractors. We project the query and candidates into the same $k=16$ space, compute cosine distance from the query to each candidate, and rank candidates from nearest to farthest. 
We compare \mn{} with the original representation and dimension-matched
Random, PCA, cPCA, and \cn{} subspaces.
Pairwise accuracy measures whether the SP candidate is ranked closer than an SC distractor, MRR measures the reciprocal rank of the SP candidate, and Hit@1 records whether the SP candidate is ranked first. 

\vspace{-0.5em}
Fisher-IRG achieves the highest MRR for all seven models and the highest pairwise accuracy and Hit@1 for six of seven. Averaged across architectures, it improves pairwise accuracy from $0.949$ to $0.959$, MRR from $0.921$ to $0.931$, and Hit@1 from $0.873$ to $0.885$ relative to the original representation. It also exceeds Random and Cov-IRG in pairwise accuracy and MRR for every model, showing that the gain is not explained by dimensionality reduction alone. 
Together, these results show that Fisher-IRG generalizes beyond discovery neighborhoods, preserving SP/SC structure on held-out perturbations while supporting direct semantic retrieval. 

%%%
%%%
%%%

%Bibliography
% \newpage
\bibliographystyle{plainnat}  
\bibliography{references}  

\newpage
\appendix
\section*{Appendix}

\startcontents[appendix]
\section*{Appendix Contents}
\printcontents[appendix]{}{1}{}
% \newpage

\vspace{-0.25em}
\section{Dataset Construction and Model Details}
\label{app:data}
\vspace{-0.25em}

\vspace{-0.5em}
\paragraph{Text dataset construction.}
The text dataset is organized into 1,800 anchor-centered semantic neighborhoods. Each neighborhood contains one anchor prompt, 10 semantic-preserving (SP) variants, and 64 semantic-changing (SC) variants, yielding 135,000 sentences in total. The purpose of this construction is to control two qualitatively different sources of local variation around the same semantic reference point. SP variants preserve the propositional content of the anchor while modifying its surface realization through changes in lexical choice, syntax, or phrasing. SC variants instead modify the semantic content while retaining sufficient structural similarity to the anchor to support a controlled local comparison.

The 64 SC variants are divided equally into 32 easy and 32 hard cases. Easy SC variants introduce
explicit changes to one or a small number of semantic attributes, producing a clearly identifiable
semantic displacement. Hard SC variants introduce broader semantic changes involving multiple
attributes or relations, producing a more substantial departure from the anchor's meaning. 
This graded construction prevents the SP--SC distinction from reducing to a comparison between paraphrases and arbitrarily unrelated sentences. An example neighborhood is

\begin{table}[h]
\centering
\small
\setlength{\tabcolsep}{5pt}
\renewcommand{\arraystretch}{1.15}
\begin{tabular}{p{0.15\columnwidth}p{0.77\columnwidth}}
\toprule
\textbf{Type} & \textbf{Text} \\
\midrule

Anchor
& The city council approved the proposal to build a new public library
downtown. \\

SP-1
& The proposal for constructing a new public library downtown was approved
by the city council. \\

SP-2
& City council members authorized the plan to construct a new downtown
public library. \\

SP-3
& The council approved plans for a new public library in the downtown area. \\

Easy SC
& The city council rejected the proposal to build a new public library
downtown. \\

Hard SC
& The city council approved the proposal to renovate an existing private
library outside the downtown area. \\

\bottomrule
\end{tabular}
\caption{
Example text neighborhood containing an anchor, representative
semantic-preserving variants, and easy and hard semantic-changing variants.
The SP variants alter wording and sentence structure while preserving the
anchor's meaning. The easy SC variant introduces an explicit change in the
council's decision, whereas the hard SC variant retains much of the anchor's
surface form while changing several related semantic attributes.
}
\label{tab:text_neighborhood_example}
\end{table}

The available SP and SC samples are divided into disjoint discovery and evaluation subsets. Samples in the discovery subset are used to construct the neighborhood-level Fisher and covariance operators and recover their corresponding subspaces. Evaluation variants are never used during subspace estimation and are reserved for the held-out representation, intervention, and semantic-matching analyses. This separation is important because the objective is not merely to characterize the perturbations from which a subspace is estimated, but to test whether the recovered geometry transfers to unseen variations from the same controlled semantic neighborhood. 
The exact held-out subsets used by each evaluation follow the corresponding experimental protocol.

\vspace{-0.5em}
\paragraph{Vision dataset construction.}
The vision dataset is constructed from 1,000 source images selected from Visual Genome. The object-level annotations and relational descriptions provided by Visual Genome allow semantic changes to be controlled at the level of objects, attributes, relations, and contextual information. 
Each source example defines one anchor-centered neighborhood containing
10 SP variants and 64 SC variants, yielding 75,000 image--description instances in total.
The SC variants span easy, medium, and hard semantic-change conditions.

\begin{figure*}[ht]
    \centering
    \includegraphics[width=\textwidth]{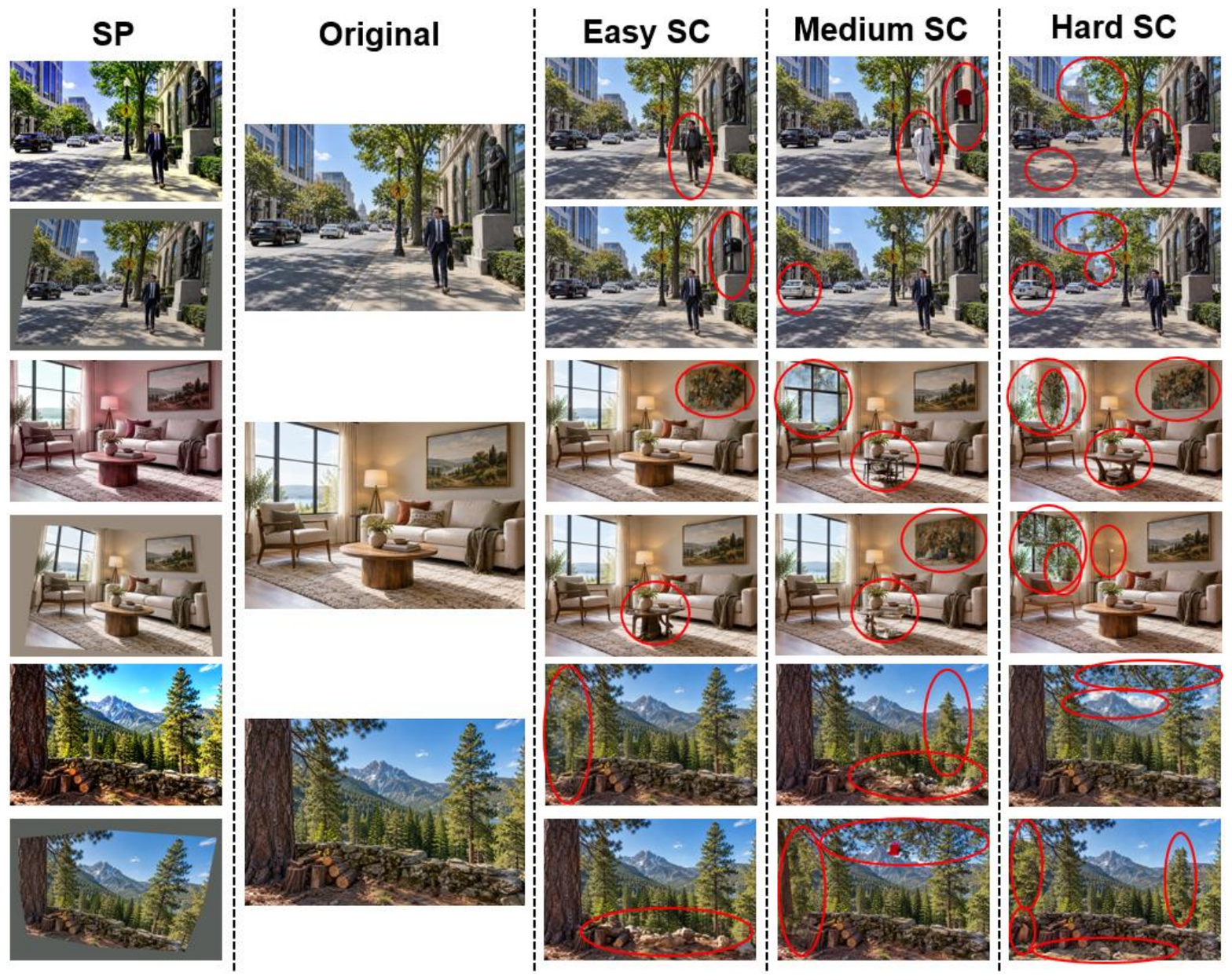}
    \caption{Additional examples of visual semantic neighborhoods.
    Each source image is paired with SP variants and SC variants at easy, medium, and hard difficulty levels. Red circles manually highlight the modified regions for readability.}
    \label{fig:vision_examples}
\end{figure*}

SP variants preserve the principal semantic content of the source example, including its relevant objects, attributes, and relations, while modifying nuisance factors without changing the underlying semantic interpretation. The SC variants span easy, medium, and hard semantic-change conditions. Easy SC variants modify one or two clearly distinguishable objects or attributes. Medium SC variants modify approximately three or four semantic elements and may additionally alter contextual information. Hard SC variants introduce multiple interacting semantic modifications or replace elements with semantically similar alternatives, creating more difficult semantic distinctions. This graded construction allows us to evaluate whether the discovered geometry remains sensitive not only to large semantic changes but also to comparatively subtle modifications.
Fig.~\ref{fig:vision_examples} provides additional examples of the visual neighborhood construction across multiple source images. SP variants preserve the source semantics, while easy, medium, and hard SC variants introduce progressively stronger semantic modifications; changed regions are manually highlighted with red circles for readability.

For held-out visual evaluation, the available variants are partitioned before subspace construction. Of the 10 SP variants, seven are used for discovery and three are reserved for evaluation. Of the 64 SC variants, 50 are used for discovery and 14 are reserved for evaluation. Thus, all reported held-out distances are computed from variants that do not contribute to estimation of either the Fisher or covariance subspace.

\begin{table}[h]
\centering
\small
\vspace{-0.5em}
\caption{Composition of the controlled semantic neighborhoods used for language and vision experiments. Variant counts are reported per anchor.}
\label{tab:appendix_dataset}
% \begin{tabular}{lccccc}
% \toprule
% Modality & Anchors & SP & Easy SC & Medium SC & Hard SC \\
% \midrule
% Text & 1,800 & 10 & 5 & -- & 5 \\
% Vision & 1,000 & 5 & 5 & 5 & 5 \\
% \bottomrule
% \end{tabular}
\begin{tabular}{lrrrr}
\toprule
Modality & Anchors & SP & SC & Total / anchor \\
\midrule
Text   & 1,800 & 10 & 64 & 75 \\
Vision & 1,000 & 10 & 64 & 75 \\
\bottomrule
\end{tabular}
\vspace{-1.0em}
\end{table}

\vspace{-0.5em}
\paragraph{Language models.}
We evaluate seven pretrained language-model families: Mistral-Instruct-v0.3, LLaMA-3-Instruct, Gemma-IT, Qwen2.5-Instruct, DeepSeek-MoE-Chat, Phi-4-Mini, and Falcon3-Instruct. The models span approximately 4B--16B parameters and were selected to provide variation in model family, scale, and training design rather than multiple checkpoints from a single architecture. In particular, the collection includes models with different pretraining and post-training pipelines as well as both dense and mixture-of-experts architectures. We analyze every model independently in its native representation space; no alignment between hidden dimensions of different models is required.

For the language experiments, hidden states are extracted from the evaluated transformer layers and aggregated using mean pooling unless otherwise specified. The primary experiments use a subspace dimension of $k=16$, while sensitivity to $k\in\{8,16,32\}$ and the effect of alternative pooling strategies are evaluated separately. The predictive readout used to construct the Fisher operator is defined from the frozen language model; additional details on the readout, Fisher approximation, and intervention implementation are provided in Appendix~\ref{app:implementation}. 

\vspace{-0.5em}
\paragraph{Vision models.}
We evaluate three transformer-based visual encoders: ViT-B/16, DINOv2 ViT-B/14, and CLIP ViT-B/16. Although the models have comparable backbone scale, they differ substantially in how their representations are learned. ViT-B/16 represents supervised visual recognition, DINOv2 ViT-B/14 represents self-supervised representation learning, and CLIP ViT-B/16 represents image--text contrastive pretraining. This design provides a controlled comparison across substantially different learning objectives while retaining broadly comparable transformer-based visual representations.

Vision experiments use the CLS-token representation at the evaluated layer. As in the language experiments, the default invariant-subspace dimension is $k=16$, with additional dimensions evaluated in the stability analysis. The predictive head, candidate set, temperature, and intervention convention used to define the Fisher geometry are fixed within each experiment and are described in Appendix~\ref{app:implementation}.

\begin{table}[h]
\centering
\small
\vspace{-1.5em}
\caption{Language and vision models used in the experiments. The suite spans different model families, scales, architectures, and representation-learning objectives.}
\label{tab:appendix_models}
\begin{tabular}{lll}
\toprule
Modality & Model & Scale / training characteristic \\
\midrule
Text & Mistral-Instruct-v0.3 & 7B \\
Text & LLaMA-3-Instruct & 8B \\
Text & Gemma-IT & 7B \\
Text & Qwen2.5-Instruct & 7B \\
Text & DeepSeek-MoE-Chat & 16B, mixture-of-experts \\
Text & Phi-4-Mini & 4B \\
Text & Falcon3-Instruct & 7B \\
\midrule
Vision & ViT-B/16 & 86M, supervised \\
Vision & DINOv2 ViT-B/14 & 86M, self-supervised \\
Vision & CLIP ViT-B/16 & 86M, image--text contrastive \\
\bottomrule
\end{tabular}
\vspace{-1.0em}
\end{table}

\vspace{-0.5em}
\paragraph{Cross-modality evaluation protocol.}
The language and vision experiments follow the same conceptual protocol despite differences in their predictive readouts. Each anchor defines a controlled local neighborhood in which SP samples instantiate meaning-preserving variation and SC samples instantiate semantic variation. The proposed Fisher-IRG and the covariance-based invariant geometry baseline, denoted Cov-IRG, are estimated from the same discovery samples at matched layers and subspace dimensions, ensuring that differences between the two methods arise from the underlying geometric operator rather than from different data. Evaluation is then performed on disjoint samples whenever the experiment tests semantic generalization. The two modalities are therefore used as independent tests of the same geometric hypothesis rather than as directly comparable numerical benchmarks; absolute predictive divergences and representation distances need not be comparable across language and vision models.

%%%
%%%
%%% Delta z vs D_kl(delta z)

\section{Additional Results for Displacement--Consequence Separability}
\label{app:euclidean_predictive}

This section provides the complete layer-wise results corresponding to Section~\ref{sec:euclidean_predictive}. The purpose of this analysis is not to test whether Euclidean displacement and predictive consequence are globally independent. Rather, we ask whether there exist local representation regimes in which perturbations of nearly equal Euclidean magnitude induce substantially different changes in the model's predictive distribution.

For an anchor representation $z$ and a held-out perturbation $z'=z+\Delta z$, we distinguish the Euclidean displacement magnitude
\[
r=\|\Delta z\|_2
\]
from its predictive consequence
\[
D_{\mathrm{KL}}\!\left(p(z)\,\|\,p(z+\Delta z)\right).
\]
The former measures how far the representation moves in the ambient Euclidean geometry, whereas the latter measures how strongly that movement changes the model's predictive distribution. Locally, the two are related through the Fisher quadratic form,
\[
D_{\mathrm{KL}}\!\left(p(z)\,\|\,p(z+\Delta z)\right)
=
\frac{1}{2}\Delta z^\top F(z)\Delta z
+
o(\|\Delta z\|_2^2).
\]
Thus, perturbations of comparable Euclidean magnitude can have substantially different predictive consequences when they are oriented along directions with different local Fisher sensitivity.

For each semantic group and layer, we consider the held-out SP and SC perturbations and select the SP--SC pair whose Euclidean displacement magnitudes are most closely matched. Writing
\[
r_{\mathrm{SP}}
=
\|\Delta z_{\mathrm{SP}}\|_2,
\qquad
r_{\mathrm{SC}}
=
\|\Delta z_{\mathrm{SC}}\|_2,
\]
we quantify their relative displacement mismatch by
\[
\delta_{\mathrm{L2}}
=
\frac{
|r_{\mathrm{SC}}-r_{\mathrm{SP}}|
}{
\max(r_{\mathrm{SC}},r_{\mathrm{SP}})
}.
\]
A semantic group is counted as \emph{Matched} when its closest SP--SC pair satisfies
\[
\delta_{\mathrm{L2}}\leq 0.10.
\]
This criterion requires the two perturbations to differ in Euclidean displacement magnitude by at most $10\%$ relative to the larger displacement.

For each matched pair, we then compare predictive consequences through
\[
\rho_{\mathrm{KL}}
=
\frac{
D_{\mathrm{KL}}^{\mathrm{SC}}
}{
D_{\mathrm{KL}}^{\mathrm{SP}}
},
\]
where $D_{\mathrm{KL}}^{\mathrm{SC}}$ and $D_{\mathrm{KL}}^{\mathrm{SP}}$ denote the predictive KL divergences produced by the matched SC and SP perturbations, respectively. A matched semantic group is counted as \emph{Qualifying} when
\[
\rho_{\mathrm{KL}}\geq 2,
\]
i.e., the SC perturbation changes the predictive distribution by at least twice as much as the nearly distance-matched SP perturbation.

The columns reported below therefore have the following precise meanings. \emph{Matched} is the number of semantic groups containing an SP--SC pair satisfying $\delta_{\mathrm{L2}}\leq0.10$. \emph{Qualifying} is the subset of those matched groups additionally satisfying $\rho_{\mathrm{KL}}\geq2$. \emph{Fraction} is the prevalence of predictive-consequence mismatches among Euclidean-matched groups,
\[
\mathrm{Fraction}
=
\frac{\mathrm{Qualifying}}{\mathrm{Matched}}.
\]
\emph{Median L2 Gap} is the median value of $\delta_{\mathrm{L2}}$ over the qualifying groups only and is reported as a percentage. It therefore measures how closely the SP and SC perturbations are Euclidean-matched among the cases that exhibit a predictive mismatch. \emph{Median KL Ratio} is the median value of $\rho_{\mathrm{KL}}$ over the same qualifying groups and measures the typical multiplicative increase in predictive change for SC relative to its distance-matched SP counterpart. When no group qualifies at a layer, these two medians are undefined and are reported as ``--''.

The threshold $\rho_{\mathrm{KL}}\geq2$ defines the mismatch event itself. Consequently, the reported Median KL Ratio should be interpreted descriptively rather than as an independent statistical test. The primary evidence is the prevalence of qualifying cases under tight Euclidean matching and their recurrence across architectures, layers, and modalities.

\subsection{Language Models}

Table~\ref{tab:appendix_language_decoupling} reports the complete language-model results, grouped by architecture. The phenomenon is strongly layer dependent, as expected for a local representation geometry. Nevertheless, qualifying cases recur across substantial portions of the networks. Mistral exhibits at least one qualifying group at every evaluated layer, Qwen2.5 at 27 of 28 layers, LLaMA-3 at 22 of 32 layers, and Gemma at all 28 evaluated layers. Thus, displacement--consequence separability is neither confined to a single depth nor universal at every representation state.

\begin{longtable}{lrrrrr}
\caption{Complete layer-wise displacement--consequence separability results for language models. \emph{Matched} counts semantic groups satisfying $\delta_{\mathrm{L2}}\leq0.10$; \emph{Qualifying} additionally requires $\rho_{\mathrm{KL}}\geq2$; \emph{Fraction} is Qualifying/Matched. Median L2 Gap and Median KL Ratio are computed only over qualifying groups.}
\label{tab:appendix_language_decoupling}\\
\toprule
Layer & Matched & Qualifying & Fraction & Median L2 Gap & Median KL Ratio \\
\midrule
\endfirsthead

\multicolumn{6}{c}{\tablename\ \thetable\ -- continued} \\
\toprule
Layer & Matched & Qualifying & Fraction & Median L2 Gap & Median KL Ratio \\
\midrule
\endhead

\midrule
\multicolumn{6}{r}{Continued on next page} \\
\endfoot

\bottomrule
\endlastfoot

\multicolumn{6}{l}{\textit{Mistral}} \\
1  & 109 & 50 & 45.87\% & 2.57\% & 16.99$\times$ \\
2  & 210 & 17 & 8.10\%  & 0.08\% & 2.40$\times$ \\
3  & 210 & 11 & 5.24\%  & 0.19\% & 2.36$\times$ \\
4  & 210 & 19 & 9.05\%  & 1.00\% & 2.23$\times$ \\
5  & 210 & 30 & 14.29\% & 3.46\% & 3.14$\times$ \\
6  & 212 & 40 & 18.87\% & 3.55\% & 2.93$\times$ \\
7  & 216 & 64 & 29.63\% & 4.17\% & 2.70$\times$ \\
8  & 220 & 74 & 33.64\% & 4.95\% & 2.84$\times$ \\
9  & 219 & 76 & 34.70\% & 4.68\% & 3.10$\times$ \\
10 & 209 & 74 & 35.41\% & 5.62\% & 3.55$\times$ \\
11 & 216 & 88 & 40.74\% & 5.17\% & 3.88$\times$ \\
12 & 200 & 84 & 42.00\% & 5.16\% & 3.72$\times$ \\
13 & 200 & 85 & 42.50\% & 5.99\% & 4.68$\times$ \\
14 & 187 & 75 & 40.11\% & 3.23\% & 4.17$\times$ \\
15 & 140 & 58 & 41.43\% & 3.29\% & 3.56$\times$ \\
16 & 161 & 82 & 50.93\% & 4.77\% & 3.33$\times$ \\
17 & 188 & 98 & 52.13\% & 5.35\% & 3.65$\times$ \\
18 & 139 & 65 & 46.76\% & 4.75\% & 3.67$\times$ \\
19 & 129 & 40 & 31.01\% & 6.72\% & 3.40$\times$ \\
20 & 145 & 54 & 37.24\% & 3.59\% & 3.35$\times$ \\
21 & 129 & 36 & 27.91\% & 3.66\% & 3.15$\times$ \\
22 & 106 & 21 & 19.81\% & 7.11\% & 3.02$\times$ \\
23 & 100 & 13 & 13.00\% & 7.90\% & 3.35$\times$ \\
24 & 96  & 5  & 5.21\%  & 9.32\% & 2.72$\times$ \\
25 & 97  & 3  & 3.09\%  & 7.43\% & 2.17$\times$ \\
26 & 91  & 3  & 3.30\%  & 5.17\% & 2.16$\times$ \\
27 & 99  & 3  & 3.03\%  & 8.94\% & 2.45$\times$ \\
28 & 104 & 1  & 0.96\%  & 6.03\% & 2.13$\times$ \\
29 & 127 & 1  & 0.79\%  & 2.87\% & 2.11$\times$ \\
30 & 147 & 2  & 1.36\%  & 3.68\% & 2.03$\times$ \\
31 & 201 & 1  & 0.50\%  & 8.99\% & 2.70$\times$ \\
32 & 392 & 33 & 8.42\%  & 4.96\% & 2.47$\times$ \\

\midrule
\multicolumn{6}{l}{\textit{Qwen2.5}} \\
1  & 646 & 71  & 10.99\% & 3.18\% & 3.07$\times$ \\
2  & 707 & 104 & 14.71\% & 3.61\% & 3.12$\times$ \\
3  & 544 & 97  & 17.83\% & 4.37\% & 3.95$\times$ \\
4  & 299 & 37  & 12.37\% & 4.10\% & 5.49$\times$ \\
5  & 239 & 1   & 0.42\%  & 7.93\% & 3.20$\times$ \\
6  & 236 & 20  & 8.47\%  & 7.22\% & 3.33$\times$ \\
7  & 233 & 0   & 0.00\%  & -- & -- \\
8  & 228 & 15  & 6.58\%  & 7.23\% & 3.86$\times$ \\
9  & 228 & 25  & 10.96\% & 6.84\% & 4.11$\times$ \\
10 & 233 & 22  & 9.44\%  & 5.83\% & 4.52$\times$ \\
11 & 236 & 23  & 9.75\%  & 6.52\% & 4.75$\times$ \\
12 & 229 & 28  & 12.23\% & 3.63\% & 4.48$\times$ \\
13 & 230 & 24  & 10.43\% & 2.88\% & 3.39$\times$ \\
14 & 228 & 26  & 11.40\% & 2.63\% & 3.52$\times$ \\
15 & 228 & 31  & 13.60\% & 2.02\% & 3.10$\times$ \\
16 & 230 & 30  & 13.04\% & 1.35\% & 3.56$\times$ \\
17 & 235 & 35  & 14.89\% & 1.64\% & 4.00$\times$ \\
18 & 229 & 29  & 12.66\% & 1.40\% & 3.18$\times$ \\
19 & 231 & 43  & 18.61\% & 1.10\% & 3.39$\times$ \\
20 & 232 & 30  & 12.93\% & 0.99\% & 3.46$\times$ \\
21 & 236 & 28  & 11.86\% & 3.07\% & 3.01$\times$ \\
22 & 247 & 35  & 14.17\% & 5.14\% & 2.84$\times$ \\
23 & 253 & 34  & 13.44\% & 6.88\% & 3.09$\times$ \\
24 & 249 & 28  & 11.24\% & 5.90\% & 3.60$\times$ \\
25 & 265 & 42  & 15.85\% & 4.94\% & 7.15$\times$ \\
26 & 267 & 29  & 10.86\% & 6.16\% & 7.57$\times$ \\
27 & 258 & 32  & 12.40\% & 6.44\% & 5.91$\times$ \\
28 & 497 & 112 & 22.54\% & 5.37\% & 3.45$\times$ \\

\midrule
\multicolumn{6}{l}{\textit{LLaMA-3}} \\
1  & 111 & 30 & 27.03\% & 8.38\% & 2.71$\times$ \\
2  & 205 & 0  & 0.00\%  & -- & -- \\
3  & 205 & 5  & 2.44\%  & 2.48\% & 2.03$\times$ \\
4  & 205 & 8  & 3.90\%  & 1.67\% & 2.14$\times$ \\
5  & 205 & 20 & 9.76\%  & 2.23\% & 2.41$\times$ \\
6  & 199 & 27 & 13.57\% & 5.53\% & 2.77$\times$ \\
7  & 191 & 14 & 7.33\%  & 6.54\% & 2.51$\times$ \\
8  & 191 & 6  & 3.14\%  & 8.17\% & 2.54$\times$ \\
9  & 194 & 6  & 3.09\%  & 7.74\% & 2.20$\times$ \\
10 & 194 & 6  & 3.09\%  & 6.70\% & 2.31$\times$ \\
11 & 180 & 4  & 2.22\%  & 8.39\% & 2.41$\times$ \\
12 & 182 & 9  & 4.95\%  & 8.18\% & 2.35$\times$ \\
13 & 172 & 15 & 8.72\%  & 7.34\% & 2.16$\times$ \\
14 & 161 & 5  & 3.11\%  & 8.38\% & 2.16$\times$ \\
15 & 153 & 21 & 13.73\% & 6.55\% & 2.17$\times$ \\
16 & 181 & 27 & 14.92\% & 9.08\% & 2.47$\times$ \\
17 & 168 & 23 & 13.69\% & 3.55\% & 2.51$\times$ \\
18 & 159 & 20 & 12.58\% & 1.37\% & 2.36$\times$ \\
19 & 166 & 7  & 4.22\%  & 6.32\% & 2.14$\times$ \\
20 & 166 & 8  & 4.82\%  & 8.02\% & 2.31$\times$ \\
21 & 161 & 0  & 0.00\%  & -- & -- \\
22 & 159 & 0  & 0.00\%  & -- & -- \\
23 & 134 & 0  & 0.00\%  & -- & -- \\
24 & 121 & 0  & 0.00\%  & -- & -- \\
25 & 115 & 0  & 0.00\%  & -- & -- \\
26 & 95  & 0  & 0.00\%  & -- & -- \\
27 & 83  & 0  & 0.00\%  & -- & -- \\
28 & 79  & 0  & 0.00\%  & -- & -- \\
29 & 89  & 0  & 0.00\%  & -- & -- \\
30 & 96  & 3  & 3.12\%  & 7.05\% & 4.27$\times$ \\
31 & 99  & 6  & 6.06\%  & 4.91\% & 2.43$\times$ \\
32 & 131 & 18 & 13.74\% & 6.92\% & 2.69$\times$ \\

\midrule
\multicolumn{6}{l}{\textit{Gemma}} \\
1  & 85  & 19 & 22.35\% & 3.48\% & 3.35$\times$ \\
2  & 96  & 23 & 23.96\% & 1.57\% & 3.08$\times$ \\
3  & 125 & 31 & 24.80\% & 3.55\% & 3.44$\times$ \\
4  & 126 & 24 & 19.05\% & 5.19\% & 2.40$\times$ \\
5  & 139 & 11 & 7.91\%  & 5.67\% & 5.51$\times$ \\
6  & 148 & 38 & 25.68\% & 6.04\% & 2.75$\times$ \\
7  & 116 & 36 & 31.03\% & 3.47\% & 4.09$\times$ \\
8  & 117 & 20 & 17.09\% & 7.00\% & 8.38$\times$ \\
9  & 132 & 13 & 9.85\%  & 7.13\% & 5.25$\times$ \\
10 & 135 & 20 & 14.81\% & 8.01\% & 3.10$\times$ \\
11 & 122 & 11 & 9.02\%  & 6.25\% & 10.04$\times$ \\
12 & 125 & 14 & 11.20\% & 4.95\% & 10.91$\times$ \\
13 & 126 & 14 & 11.11\% & 5.91\% & 6.51$\times$ \\
14 & 124 & 13 & 10.48\% & 5.52\% & 4.94$\times$ \\
15 & 127 & 7  & 5.51\%  & 4.03\% & 6.37$\times$ \\
16 & 131 & 23 & 17.56\% & 7.71\% & 5.25$\times$ \\
17 & 135 & 35 & 25.93\% & 7.86\% & 3.98$\times$ \\
18 & 119 & 32 & 26.89\% & 5.80\% & 3.16$\times$ \\
19 & 112 & 27 & 24.11\% & 4.74\% & 3.64$\times$ \\
20 & 94  & 15 & 15.96\% & 6.58\% & 8.01$\times$ \\
21 & 107 & 8  & 7.48\%  & 1.01\% & 7.71$\times$ \\
22 & 120 & 15 & 12.50\% & 5.90\% & 4.27$\times$ \\
23 & 134 & 16 & 11.94\% & 4.13\% & 3.71$\times$ \\
24 & 135 & 31 & 22.96\% & 4.40\% & 3.34$\times$ \\
25 & 112 & 16 & 14.29\% & 4.02\% & 3.16$\times$ \\
26 & 80  & 6  & 7.50\%  & 3.66\% & 187.50$\times$ \\
27 & 85  & 8  & 9.41\%  & 6.03\% & 4.49$\times$ \\
28 & 198 & 56 & 28.28\% & 4.31\% & 3.17$\times$ \\

\end{longtable}

The unusually large Median KL Ratio for Gemma at layer 26 arises from a very small matched SP predictive divergence in the denominator of $\rho_{\mathrm{KL}}$. It should therefore not be interpreted as evidence that this layer is uniquely separated by two orders of magnitude. The prevalence of qualifying distance-matched cases, rather than the magnitude of this extreme ratio, is the primary quantity of interest.

\subsection{Vision Model}

Table~\ref{tab:appendix_vit_decoupling} reports the complete ViT-B/16 results. Predictive-consequence mismatches occur at every evaluated layer. Euclidean matching is particularly tight: among qualifying groups, the layer-wise Median L2 Gap ranges from $1.28\%$ to $3.14\%$, whereas the Median KL Ratio ranges from $2.43\times$ to $3.59\times$. Thus, even when SP and SC perturbations induce nearly equal representation displacement, their predictive consequences can differ substantially. This provides a cross-modal counterpart to the language results and shows that displacement--consequence separability is not specific to autoregressive language representations.

\begin{table}[ht]
\centering
\small
\setlength{\tabcolsep}{4.5pt}
\renewcommand{\arraystretch}{1.06}
\begin{tabular}{rrrrrr}
\toprule
Layer & Matched & Qualifying & Fraction & Median L2 Gap & Median KL Ratio \\
\midrule
1  & 348 & 72 & 20.69\% & 3.14\% & 3.04$\times$ \\
2  & 410 & 96 & 23.41\% & 2.25\% & 3.20$\times$ \\
3  & 406 & 32 & 7.88\%  & 1.77\% & 2.72$\times$ \\
4  & 409 & 16 & 3.91\%  & 2.38\% & 2.72$\times$ \\
5  & 416 & 26 & 6.25\%  & 1.65\% & 2.75$\times$ \\
6  & 404 & 20 & 4.95\%  & 1.68\% & 2.98$\times$ \\
7  & 401 & 27 & 6.73\%  & 2.50\% & 2.43$\times$ \\
8  & 339 & 39 & 11.50\% & 2.54\% & 2.86$\times$ \\
9  & 308 & 42 & 13.64\% & 1.28\% & 2.62$\times$ \\
10 & 311 & 41 & 13.18\% & 2.53\% & 3.25$\times$ \\
11 & 308 & 37 & 12.01\% & 1.97\% & 3.59$\times$ \\
12 & 308 & 39 & 12.66\% & 1.75\% & 2.95$\times$ \\
\bottomrule
\end{tabular}
\caption{Complete layer-wise displacement--consequence separability results for ViT-B/16 using CLS-token representations. Column definitions are identical to Table~\ref{tab:appendix_language_decoupling}.}
\label{tab:appendix_vit_decoupling}
\end{table}

Across language and vision, these results establish a limited but important geometric claim. Euclidean displacement magnitude and predictive consequence may be related, but Euclidean magnitude alone is insufficient to determine the behavioral significance of a local representation change. Distance-matched counterexamples recur across architectures, depths, and modalities, supporting the need for the predictive geometry used by \mn\ without requiring the stronger claim that Euclidean and predictive geometries are globally decoupled.

%%%
%%%
%%%

%%%
%%%
%%% layer prediction

\section{Full Layer-wise Predictive Selectivity Results}
\label{app:localization}
\vspace{-0.5em}

This section provides the complete layer-wise comparison corresponding to Section~\ref{sec:selectivity_stability}. The purpose of the experiment is to determine whether the local geometry induced by Fisher predictive sensitivity yields stronger semantic--nuisance separation than the covariance-based counterpart across model depth.

\vspace{-0.5em}
\paragraph{Evaluation protocol.}
For each semantic group $g$, layer $\ell$, and geometry $m\in\{\mathrm{Fisher},\mathrm{Cov}\}$, we estimate the corresponding invariant subspace from the discovery variants and evaluate its predictive effect using semantic-preserving (SP) and semantic-changing (SC) interventions. Let
$D^{m,\mathrm{SP}}_{g,\ell}$ and
$D^{m,\mathrm{SC}}_{g,\ell}$
denote the output divergences induced by the two intervention types. We define
\[
S^{m}_{g,\ell}
=
\log
\frac{
D^{m,\mathrm{SC}}_{g,\ell}+\delta
}{
D^{m,\mathrm{SP}}_{g,\ell}+\delta
},
\]
where $\delta>0$ is a small numerical stabilizer used only in the selectivity ratio. 
We compare Fisher-IRG with Cov-IRG through
\[
\Delta S_{g,\ell}
=
S^{\mathrm{Fisher}}_{g,\ell}
-
S^{\mathrm{Cov}}_{g,\ell}.
\]
Positive $\Delta S$ indicates stronger semantic-versus-nuisance predictive selectivity under Fisher-IRG. Reported values are first computed at the semantic-group level and then averaged across groups.

Because the evaluated architectures contain different numbers of layers, we express depth in normalized coordinates $\ell/L$ and report five approximately evenly spaced locations. Language experiments use mean-pooled hidden representations, vision experiments use CLS-token representations, and the default subspace dimension is $k=16$.

\vspace{-0.5em}
\paragraph{Why uniform Fisher dominance is not expected.}
Fisher-IRG is not expected to outperform Cov-IRG at every possible layer, semantic group, or local neighborhood. The Fisher metric measures the local effect of representation perturbations on the model's predictive distribution, and this sensitivity is itself representation dependent. As information is progressively transformed across depth, different layers can vary substantially in how strongly local directions are coupled to the downstream prediction. Some layers may primarily encode intermediate or weakly prediction-aligned features, whereas others may contain directions whose perturbation has a much larger behavioral consequence. Consequently, the relative advantage of predictive weighting can vary with both architecture and depth.

Our hypothesis is therefore not that Fisher weighting produces a uniformly larger selectivity score at every local comparison. Rather, it predicts a systematic advantage when predictive sensitivity is an informative geometric quantity for the representation under analysis. The relevant evidence is thus consistency of the aggregate trend across architectures, modalities, and depth, together with model-specific localization of the strongest effect. This distinction is important because it separates the geometric claim from an unrealistically uniform dominance claim.

\vspace{-0.5em}
\paragraph{Complete layer-wise comparison.}
Table~\ref{tab:full_selectivity} reports the full normalized-depth results. Fisher-IRG yields positive mean $\Delta S$ at every displayed depth for all seven language models and all three vision models. However, the magnitude of this advantage is clearly architecture dependent.

\begin{table}[ht]
\centering
\small
\vspace{-0.5em}
\caption{Full layer-wise predictive-selectivity advantage of Fisher-IRG over Cov-IRG. Each entry reports $\Delta S=S^{\mathrm{Fisher}}-S^{\mathrm{Cov}}$, averaged across semantic groups. Positive values indicate stronger semantic-versus-nuisance predictive selectivity under Fisher-IRG.}
\label{tab:full_selectivity}
\setlength{\tabcolsep}{4.0pt}
\begin{tabular}{lccccc}
\toprule
Model & $0.00$ & $0.25$ & $0.50$ & $0.75$ & $1.00$ \\
\midrule
\multicolumn{6}{l}{\textit{Language}} \\
LLaMA-3-Instruct      & 0.140 & 0.270 & 0.370 & 0.190 & 0.120 \\
Mistral-Instruct-v0.3 & 0.120 & 0.330 & 0.170 & 0.220 & 0.310 \\
Gemma-IT              & 0.260 & 0.170 & 0.120 & 0.090 & 0.210 \\
Qwen2.5-Instruct      & 0.280 & 0.340 & 0.470 & 0.520 & 0.230 \\
DeepSeek-MoE-Chat     & 0.220 & 0.410 & 0.180 & 0.240 & 0.190 \\
Phi-4-Mini            & 0.110 & 0.360 & 0.300 & 0.230 & 0.110 \\
Falcon3-Instruct      & 0.340 & 0.240 & 0.170 & 0.190 & 0.230 \\
\midrule
\multicolumn{6}{l}{\textit{Vision}} \\
ViT-B/16              & 0.720 & 0.690 & 0.480 & 0.660 & 0.590 \\
DINOv2 ViT-B/14       & 0.570 & 0.450 & 0.550 & 0.540 & 0.470 \\
CLIP ViT-B/16         & 0.610 & 0.430 & 0.440 & 0.590 & 0.310 \\
\bottomrule
\end{tabular}
\vspace{-1.5em}
\end{table}

\vspace{-0.5em}
\paragraph{Language models.}
The language models exhibit a consistent Fisher advantage but differ in where that advantage is strongest. LLaMA-3-Instruct reaches its largest improvement near normalized depth $0.50$, whereas Mistral-Instruct-v0.3, DeepSeek-MoE-Chat, and Phi-4-Mini peak earlier, near depth $0.25$. Qwen2.5-Instruct reaches its maximum at depth $0.75$, while Gemma-IT and Falcon3-Instruct show their largest gains near the earliest evaluated depth.

This variation is consistent with the representation-dependent nature of predictive geometry. The results do not indicate a universal ``Fisher layer.'' Instead, they suggest that different architectures localize prediction-sensitive semantic structure at different depths. Importantly, despite this variation in localization, the aggregate advantage remains positive across all five reported depths for every language model.

\vspace{-0.5em}
\paragraph{Vision models.}
The same pattern appears in vision. ViT-B/16, DINOv2 ViT-B/14, and CLIP ViT-B/16 all exhibit positive $\Delta S$ across the full set of reported depths. The magnitude of the improvement is larger on average than in the language models, although these absolute values should not be compared directly across modalities because the predictive readouts and intervention mechanisms differ.

All three vision models show strong Fisher advantages in early and intermediate representations, while positive margins persist through later layers. This consistency across supervised, self-supervised, and image--text contrastive training objectives indicates that the benefit of Fisher weighting is not tied to one particular visual learning paradigm.

\vspace{-0.5em}
\paragraph{Interpretation.}
Taken together, the full results support two complementary conclusions. First, Fisher-IRG provides a systematic improvement in semantic-versus-nuisance predictive selectivity relative to Cov-IRG across model families and modalities. Second, the strength of this improvement is localized differently across architectures, consistent with the fact that predictive sensitivity changes as representations evolve through the network. The central claim is therefore one of robust cross-model advantage with architecture-dependent localization, rather than uniform layer-wise dominance as a theoretical requirement.

%%%
%%%
%%%

%%%
%%%
%%% Grassmann

\vspace{-1.0em}
\section{Additional Analysis of Invariant-Subspace Reproducibility}
\label{app:grassmann_stability}
\vspace{-0.5em}

This section provides the complete reproducibility analysis corresponding to Section~\ref{sec:selectivity_stability}. We evaluate whether independently sampled discovery neighborhoods recover similar invariant subspaces, and whether Fisher-IRG provides more reproducible geometry than Cov-IRG across model depth and subspace dimension.

\vspace{-0.5em}
\paragraph{Resampling protocol.}
For each model, evaluated layer, and subspace dimension $k$, we independently construct invariant subspaces from different sampled subsets of the available discovery variants. The resulting subspaces are compared using the Grassmann distance $d_{\mathrm{Gr}}$ defined in Section~\ref{sec:selectivity_stability}. Distances are first computed for individual semantic groups and independent run pairs and are then averaged across groups.

Fisher-IRG and Cov-IRG are evaluated under matched conditions: they use the same model representation, semantic neighborhoods, layer, subspace dimension, and resampling protocol. We consider
$k\in\{8,16,32\}$ and three representative transformer layers, $3$, $8$, and $10$. Because Grassmann distance depends on $k$, the meaningful comparison is Fisher-IRG versus Cov-IRG at the same dimension and layer; distances should not be interpreted as directly comparable across different values of $k$.

\vspace{-0.5em}
\paragraph{Full language-model results.}
Table~\ref{tab:app_grassmann_text} reports the complete language results. Fisher-IRG obtains lower Grassmann distance in 58 of the 63 matched comparisons. 
The advantage is especially consistent for DeepSeek-MoE-Chat and Falcon3-Instruct, where Fisher-IRG is more reproducible in every reported layer--dimension configuration.

\begin{table}[t]
\centering
\small
\caption{Full Grassmann reproducibility comparison for language models. Each entry reports Fisher-IRG/Cov-IRG Grassmann distance, averaged across semantic groups and independent run pairs. Lower is better.}
\label{tab:app_grassmann_text}
\setlength{\tabcolsep}{3.8pt}
\begin{tabular}{lc|ccc}
\toprule
Model & $k$ & Layer 3 & Layer 8 & Layer 10 \\
\midrule
\multirow{3}{*}{Mistral-Instruct-v0.3}
& 8  & 2.427/2.593 & 2.381/2.557 & 2.439/2.603 \\
& 16 & 2.404/2.612 & 2.639/2.763 & 2.480/2.538 \\
& 32 & 2.352/2.396 & 2.307/2.447 & 2.699/2.307 \\
\midrule
\multirow{3}{*}{LLaMA-3-Instruct}
& 8  & 2.422/2.615 & 2.685/2.715 & 2.606/2.767 \\
& 16 & 2.556/2.629 & 2.426/2.536 & 2.560/2.583 \\
& 32 & 2.621/2.594 & 2.694/2.728 & 2.411/2.626 \\
\midrule
\multirow{3}{*}{Gemma-IT}
& 8  & 2.623/2.684 & 2.878/2.807 & 2.799/2.807 \\
& 16 & 2.782/2.846 & 2.824/2.965 & 2.506/2.541 \\
& 32 & 2.648/2.751 & 2.751/2.789 & 2.777/2.842 \\
\midrule
\multirow{3}{*}{Qwen2.5-Instruct}
& 8  & 2.716/2.890 & 2.817/2.839 & 2.738/2.632 \\
& 16 & 2.638/2.941 & 2.559/2.888 & 2.678/2.787 \\
& 32 & 2.630/2.848 & 2.627/2.748 & 2.645/2.729 \\
\midrule
\multirow{3}{*}{DeepSeek-MoE-Chat}
& 8  & 2.252/2.305 & 2.274/2.323 & 2.289/2.501 \\
& 16 & 2.269/2.448 & 2.543/2.560 & 2.365/2.544 \\
& 32 & 2.211/2.321 & 2.359/2.555 & 2.413/2.548 \\
\midrule
\multirow{3}{*}{Phi-4-Mini}
& 8  & 2.380/2.401 & 2.260/2.273 & 2.205/2.548 \\
& 16 & 2.477/2.553 & 2.402/2.503 & 2.329/2.364 \\
& 32 & 2.352/2.295 & 2.236/2.459 & 2.391/2.447 \\
\midrule
\multirow{3}{*}{Falcon3-Instruct}
& 8  & 2.863/2.962 & 3.078/3.137 & 2.803/2.944 \\
& 16 & 3.110/3.199 & 3.033/3.136 & 2.962/3.107 \\
& 32 & 2.814/2.922 & 3.098/3.150 & 2.902/2.947 \\
\bottomrule
\end{tabular}
\end{table}

The remaining reversals are sparse and occur in isolated layer--dimension configurations across several architectures, rather than forming a consistent architecture-, layer-, or dimension-specific pattern.  
We do not view these reversals as inconsistent with the proposed geometry. Reproducibility depends jointly on the sampled local neighborhoods, the spectrum of the generalized eigenproblem, the effective eigengap, and the representation geometry at the analyzed layer. Fisher-IRG is therefore not expected to dominate every finite-sample estimate. The relevant empirical claim is the strong cross-model tendency toward lower Grassmann distance under predictive weighting.

\vspace{-0.5em}
\paragraph{Full vision-model results.}
Table~\ref{tab:app_grassmann_vision} reports the corresponding vision results. Fisher-IRG obtains lower Grassmann distance in 26 of the 27 evaluated layer--dimension settings. 
The only reversal occurs for CLIP ViT-B/16 at layer 3 with $k=16$, where Fisher-IRG and Cov-IRG yield Grassmann distances of 3.543 and 3.466, respectively. More broadly, Fisher-IRG remains more reproducible across supervised visual recognition, self-supervised representation learning, and image--text contrastive pretraining.

\begin{table}[t]
\centering
\small
\caption{Full Grassmann reproducibility comparison for vision models. Each entry reports Fisher-IRG/Cov-IRG Grassmann distance. All results use CLS-token representations. Lower is better.}
\label{tab:app_grassmann_vision}
\setlength{\tabcolsep}{4.5pt}
\begin{tabular}{lc|ccc}
\toprule
Model & $k$ & Layer 3 & Layer 8 & Layer 10 \\
\midrule
\multirow{3}{*}{ViT-B/16}
& 8  & 3.238/3.430 & 3.334/3.389 & 3.309/3.371 \\
& 16 & 3.329/3.405 & 3.346/3.412 & 3.199/3.254 \\
& 32 & 3.235/3.334 & 3.340/3.424 & 3.217/3.313 \\
\midrule
\multirow{3}{*}{DINOv2 ViT-B/14}
& 8  & 3.388/3.582 & 3.152/3.208 & 3.415/3.523 \\
& 16 & 3.226/3.362 & 3.598/3.791 & 3.112/3.356 \\
& 32 & 3.298/3.402 & 3.352/3.388 & 3.256/3.480 \\
\midrule
\multirow{3}{*}{CLIP ViT-B/16}
& 8  & 3.474/3.609 & 3.201/3.578 & 3.193/3.231 \\
& 16 & 3.543/3.466 & 3.162/3.350 & 3.368/3.471 \\
& 32 & 3.412/3.500 & 3.161/3.382 & 3.218/3.428 \\
\bottomrule
\end{tabular}
\end{table}

\vspace{-0.5em}
\paragraph{Dependence on layer and subspace dimension.}
Neither method exhibits a universal monotonic trend with layer depth or $k$. This is expected because increasing the dimension changes the geometric object being estimated, while different layers expose different local representation structures and predictive couplings. In particular, a larger $k$ may include directions associated with smaller generalized eigenvalues or weaker eigengaps, making the estimated subspace more sensitive to finite-sample variation. Conversely, some layers can contain a more sharply defined predictive-semantic structure and therefore yield stable estimates even at larger $k$.

Accordingly, we do not interpret the absolute Grassmann distance across different dimensions as a ranking of which $k$ is intrinsically preferable. The controlled comparison is always between Fisher-IRG and Cov-IRG at matched $k$ and layer. Under this comparison, the Fisher advantage remains broad across the tested range.

\vspace{-0.5em}
\paragraph{Subspace stability versus individual-vector stability.}
The main paper treats the recovered invariant object as a subspace rather than an ordered collection of generalized eigenvectors. As an additional diagnostic, we nevertheless examine whether reproducibility also extends to the individual vectors used to span that subspace.

Let $U^{(1)},U^{(2)}\in\mathbb{R}^{d\times k}$ denote two independently estimated orthonormal bases. We measure subspace agreement using the mean principal cosine
\[
C_{\mathrm{sub}}
=
\frac{1}{k}\sum_{i=1}^{k}
\sigma_i\!\left(
U^{(1)\top}U^{(2)}
\right),
\]
and individual-vector agreement using
\[
C_{\mathrm{vec}}
=
\frac{1}{k}
\sum_{i=1}^{k}
\left|
u_i^{(1)\top}u_i^{(2)}
\right|.
\]
The absolute value removes eigenvector sign ambiguity. We summarize the distinction through
\[
\Delta C
=
C_{\mathrm{sub}}-C_{\mathrm{vec}}.
\]
A positive $\Delta C$ indicates that the recovered span is more reproducible than a particular ordered basis within that span. This comparison is supplementary: because basis vectors can rotate, exchange ordering, or become weakly identifiable when generalized eigenvalues are close, $C_{\mathrm{vec}}$ is not the primary stability criterion used in the paper.

\begin{table}[t]
\centering
\small
\caption{Subspace-level versus individual-vector reproducibility of Fisher-IRG for language models. Results are averaged across evaluated layers, semantic groups, and independent run pairs.}
\label{tab:app_vector_stability_text}
\setlength{\tabcolsep}{5pt}
\begin{tabular}{lcccc}
\toprule
Model & $k$ & $C_{\mathrm{sub}}\uparrow$ & $C_{\mathrm{vec}}\uparrow$ & $\Delta C$ \\
\midrule
Mistral-Instruct-v0.3 & 8  & .6154 & .3636 & .2518 \\
                      & 16 & .6228 & .3458 & .2770 \\
                      & 32 & .7431 & .1489 & .5942 \\
LLaMA-3-Instruct      & 8  & .6655 & .3840 & .2815 \\
                      & 16 & .6589 & .2736 & .3853 \\
                      & 32 & .7238 & .2035 & .5203 \\
Gemma-IT              & 8  & .6413 & .3724 & .2689 \\
                      & 16 & .6551 & .3226 & .3325 \\
                      & 32 & .7067 & .2574 & .4493 \\
Qwen2.5-Instruct      & 8  & .6459 & .3705 & .2754 \\
                      & 16 & .6630 & .3856 & .2774 \\
                      & 32 & .728 & .1482 & .5798 \\
DeepSeek-MoE-Chat     & 8  & .6847 & .3604 & .3243 \\
                      & 16 & .6788 & .3521 & .3267 \\
                      & 32 & .7519 & .2892 & .4627 \\
Phi-4-Mini            & 8  & .6232 & .3481 & .2751 \\
                      & 16 & .6378 & .3111 & .3267 \\
                      & 32 & .7442 & .1905 & .5537 \\
Falcon3-Instruct      & 8  & .5927 & .3264 & .2663 \\
                      & 16 & .5730 & .3392 & .2338 \\
                      & 32 & .7193 & .2238 & .4955 \\
\bottomrule
\end{tabular}
\end{table}

Across all seven language models and all three evaluated dimensions, $C_{\mathrm{sub}}$ exceeds $C_{\mathrm{vec}}$, as expected because subspace agreement is invariant to rotations within the recovered span whereas $C_{\mathrm{vec}}$ compares a particular ordered basis. The magnitude of this gap is nevertheless informative about basis variability within a comparatively stable recovered span. Principal-cosine agreement ranges from 0.5730 to 0.7519, whereas individual-vector cosine ranges from 0.1482 to 0.3856.

The same qualitative behavior is observed for vision models across easy, medium, and hard semantic-change conditions. Across the reported configurations, principal cosine ranges from approximately $0.53$ to $0.68$, whereas individual-vector cosine ranges from approximately $0.16$ to $0.37$. We provide the complete vision breakdown in Table~\ref{tab:app_vector_stability_vision}.

\begin{table}[t]
\centering
\small
\caption{Subspace-level versus individual-vector reproducibility of Fisher-IRG for vision models across semantic-change difficulty and subspace dimension. Results use CLS-token representations and are averaged across evaluated layers, Visual Genome groups, and independent run pairs.}
\label{tab:app_vector_stability_vision}
\setlength{\tabcolsep}{3.8pt}
\begin{tabular}{lclccc}
\toprule
Model & $k$ & SC & $C_{\mathrm{sub}}\uparrow$ & $C_{\mathrm{vec}}\uparrow$ & $\Delta C$ \\
\midrule
ViT-B/16
& 8  & Easy   & .6233 & .2714 & .3519 \\
&    & Medium & .6064 & .2374 & .3690 \\
&    & Hard   & .5554 & .1936 & .3618 \\
& 16 & Easy   & .6504 & .2694 & .3810 \\
&    & Medium & .6301 & .2518 & .3783 \\
&    & Hard   & .6384 & .2603 & .3781 \\
& 32 & Easy   & .6431 & .3256 & .3175 \\
&    & Medium & .6693 & .2932 & .3761 \\
&    & Hard   & .5949 & .2480 & .3469 \\
\midrule
DINOv2 ViT-B/14
& 8  & Easy   & .6755 & .3389 & .3366 \\
&    & Medium & .6260 & .3117 & .3143 \\
&    & Hard   & .6065 & .3181 & .2884 \\
& 16 & Easy   & .6803 & .3458 & .3345 \\
&    & Medium & .6564 & .2663 & .3901 \\
&    & Hard   & .5942 & .1586 & .4356 \\
& 32 & Easy   & .6449 & .3132 & .3317 \\
&    & Medium & .6358 & .2791 & .3567 \\
&    & Hard   & .6298 & .2118 & .4180 \\
\midrule
CLIP ViT-B/16
& 8  & Easy   & .6350 & .3656 & .2694 \\
&    & Medium & .6405 & .3708 & .2697 \\
&    & Hard   & .6158 & .2939 & .3219 \\
& 16 & Easy   & .6210 & .3216 & .2994 \\
&    & Medium & .6039 & .3271 & .2768 \\
&    & Hard   & .5344 & .2992 & .2352 \\
& 32 & Easy   & .6439 & .3262 & .3177 \\
&    & Medium & .6273 & .2447 & .3826 \\
&    & Hard   & .6388 & .2096 & .4292 \\
\bottomrule
\end{tabular}
\end{table}

The positive subspace--vector gap is present for every reported vision configuration. However, neither semantic-change difficulty nor subspace dimension produces a universal monotonic trend in the gap. This is consistent with our interpretation that the primary recovered object is the invariant span: individual generalized eigenvectors may vary within that span without changing the geometry represented by the corresponding point on the Grassmann manifold.

\vspace{-0.5em}
\paragraph{Summary.}
The expanded analysis supports the main-paper conclusion at three levels. Fisher-IRG is more reproducible than Cov-IRG in the large majority of matched layer--dimension comparisons; the result persists across language and vision architectures with different training objectives; and independent Fisher estimates agree more strongly at the level of their spans than at the level of individual generalized eigenvectors. Together, these observations support treating the Fisher-induced invariant subspace, rather than any particular basis used to represent it, as the reproducible geometric object.

%%%
%%%
%%%

%%%
%%%
%%% Fisher Cov Different

\vspace{-1.0em}
\section{Additional Analysis of Fisher--Covariance Geometric Separation}
\label{app:distinct_geometry}
\vspace{-0.5em}

This section expands the direct geometric comparison in Section~\ref{sec:distinct_geometry}. The purpose of the experiment is to determine whether Fisher-IRG and Cov-IRG recover genuinely different local subspaces, rather than two noisy estimates of approximately the same representation geometry.

\vspace{-0.5em}
\paragraph{Comparison protocol.}
For each semantic group, we independently construct two Fisher-IRG subspaces,
$\mathcal{S}_{F}^{(a)}$ and $\mathcal{S}_{F}^{(b)}$, and two Cov-IRG subspaces,
$\mathcal{S}_{C}^{(a)}$ and $\mathcal{S}_{C}^{(b)}$, from independent splits of the available discovery variants. This allows us to separate variation caused by finite-sample estimation from variation caused by changing the geometric operator itself.

Let $U$ and $V$ denote orthonormal bases for two $k$-dimensional subspaces. We measure their normalized projection distance as
\[
d_{\mathrm{proj}}(U,V)
=
\frac{
\left\|
UU^\top - VV^\top
\right\|_F
}{
\sqrt{2k}
}.
\]
This quantity is basis invariant, equals zero when the two subspaces coincide, and approaches one as their spans become orthogonal. For each semantic group, we compute the within-Fisher distance
\[
d(F,F')
=
d_{\mathrm{proj}}
\left(
U_F^{(a)},U_F^{(b)}
\right),
\]
the within-covariance distance
\[
d(C,C')
=
d_{\mathrm{proj}}
\left(
U_C^{(a)},U_C^{(b)}
\right),
\]
and the cross-method distance
\[
d(F,C)
=
d_{\mathrm{proj}}
\left(
U_F^{(a)},U_C^{(a)}
\right).
\]

We summarize the separation between the two geometries using
\[
\Delta_{\mathrm{geo}}
=
d(F,C)
-
\max
\left\{
d(F,F'),
d(C,C')
\right\}.
\]
A positive $\Delta_{\mathrm{geo}}$ indicates that the Fisher--covariance discrepancy exceeds the larger source of within-method estimation variability. We additionally report the fraction of semantic groups satisfying
\[
d(F,C)
>
\max
\left\{
d(F,F'),
d(C,C')
\right\},
\]
together with the mean principal cosine between the Fisher and covariance subspaces. The principal-angle definition follows Section~\ref{sec:selectivity_stability}; smaller principal cosine indicates less overlap between the recovered spans. 

All results are computed at layer 10 with $k=16$. Language experiments use mean-pooled representations, while vision experiments use CLS-token representations.

\vspace{-0.5em}
\paragraph{Complete cross-model comparison.}
Table~\ref{tab:app_distinct_geometry} reports the full comparison. Across all seven language models and all three vision models, the Fisher--covariance distance exceeds both corresponding within-method distances for every evaluated semantic group.

\begin{table}[t]
\centering
\small
\caption{Direct geometric comparison between Fisher-IRG and Cov-IRG. $d(F,F')$ and $d(C,C')$ measure within-method variation across independent discovery splits, while $d(F,C)$ measures cross-method separation. Principal Cos. measures mean subspace overlap, and $\Delta_{\mathrm{geo}}$ is the cross-method distance minus the larger within-method distance. Positive Groups reports the fraction of semantic groups for which the cross-method distance exceeds both within-method distances.}
\label{tab:app_distinct_geometry}
\setlength{\tabcolsep}{3.6pt}
\begin{tabular}{lcccccc}
\toprule
Model
& $d(F,F')\downarrow$
& $d(C,C')\downarrow$
& $d(F,C)\uparrow$
& Prin.\ Cos.$\downarrow$
& $\Delta_{\mathrm{geo}}\uparrow$
& Pos.\ Groups \\
\midrule
\multicolumn{7}{l}{\textit{Language}} \\
Mistral-Instruct-v0.3
& .410 & .439 & .998 & .064 & .559 & 100\% \\
LLaMA-3-Instruct
& .381 & .513 & .997 & .070 & .484 & 100\% \\
Gemma-IT
& .417 & .449 & .998 & .069 & .549 & 100\% \\
Qwen2.5-Instruct
& .413 & .438 & .997 & .056 & .559 & 100\% \\
DeepSeek-MoE-Chat
& .473 & .491 & .998 & .071 & .507 & 100\% \\
Phi-4-Mini
& .394 & .412 & .993 & .069 & .581 & 100\% \\
Falcon3-Instruct
& .453 & .458 & .993 & .071 & .535 & 100\% \\
\midrule
\multicolumn{7}{l}{\textit{Vision}} \\
ViT-B/16
& .356 & .518 & .997 & .080 & .479 & 100\% \\
DINOv2 ViT-B/14
& .366 & .466 & .995 & .068 & .529 & 100\% \\
CLIP ViT-B/16
& .358 & .527 & .992 & .065 & .465 & 100\% \\
\bottomrule
\end{tabular}
\end{table}

\vspace{-0.5em}
\paragraph{Language-model geometry.}
The language models exhibit extremely limited overlap between the subspaces selected by Fisher-IRG and Cov-IRG. Across all seven architectures, the cross-method projection distance lies between $0.993$ and $0.998$, while the corresponding mean principal cosine lies between $0.056$ and $0.071$. Thus, the Fisher- and covariance-derived subspaces are nearly orthogonal under both projector-based and principal-angle measurements.

The cross-method separation also exceeds within-method variability in every case. The geometric margin $\Delta_{\mathrm{geo}}$ remains positive across all architectures, ranging from $0.484$ for LLaMa-3-Instruct to $0.581$ for Phi-4-Mini. Moreover, every evaluated semantic group satisfies the same ordering. This is important because a large average Fisher--covariance distance alone would not establish distinct geometry if independently re-estimating either method produced equally large variation. The positive group-level margins show that the difference induced by changing the geometric operator is systematically larger than the corresponding estimation variability. 

\vspace{-0.5em}
\paragraph{Vision-model geometry.}
The same qualitative pattern appears across the three vision encoders. Fisher--covariance projection distances range from $0.992$ to $0.997$, while principal cosines remain between $0.065$ and $0.080$. These values indicate greater overlap than in the language models, but the two induced subspaces remain substantially separated.

The corresponding geometric margins are positive for all three vision models: $0.479$ for ViT-B/16, $0.529$ for DINOv2 ViT-B/14, and $0.465$ for CLIP ViT-B/16. As in language, the cross-method distance exceeds both within-method distances for every evaluated semantic group. 
The result therefore persists across supervised, self-supervised, and image--text contrastive visual representations. 

\vspace{-0.5em}
\paragraph{Why geometric distinctness matters.}
The distinction between Fisher-IRG and Cov-IRG is not merely numerical. The two constructions answer different local geometric questions. Cov-IRG measures where representations vary under the controlled SP and SC neighborhoods, whereas Fisher-IRG measures how local representation directions affect the model's predictive distribution. If predictive sensitivity were only a scalar reweighting of the same dominant representation directions, the two methods could exhibit different selectivity values while still recovering strongly overlapping subspaces. The low principal cosines and large cross-method projection distances rule out this explanation empirically.

Instead, the results indicate that predictive weighting changes which directions are treated as geometrically important. In this sense, Fisher-IRG does not simply refine the covariance geometry; it induces a different local geometry on the same representation space.

\vspace{-0.5em}
\paragraph{Relation to the reproducibility analysis.}
This experiment should not be interpreted as an independent measure of absolute subspace reproducibility. In particular, the within-method projection distances can themselves be substantial, especially for the language models. Absolute reproducibility is evaluated directly using Grassmann distance in Section~\ref{sec:selectivity_stability} and Appendix~\ref{app:grassmann_stability}.

The purpose here is instead relative: we ask whether the discrepancy between Fisher-IRG and Cov-IRG is larger than the variability obtained by re-estimating either construction from independent samples. The consistently positive $\Delta_{\mathrm{geo}}$ and 100\% positive-group rate across all evaluated architectures provide that evidence.

\vspace{-0.5em}
\paragraph{Summary.}
Across language and vision models, Fisher-IRG and Cov-IRG recover subspaces whose cross-method separation consistently exceeds within-method estimation variability. The effect is especially pronounced in language, where principal-angle overlap is close to zero, but remains strong across all three vision encoders. These results support the interpretation that Fisher predictive sensitivity and covariance displacement statistics define genuinely different local geometries rather than alternative estimators of the same invariant subspace.

%%%
%%%
%%%

%%%
%%%
%%% Causal

\vspace{-0.5em}
\section{Additional Causal Validation by Representation Intervention}
\label{app:causal_intervention}
\vspace{-0.5em}

This section provides the complete group-level intervention analysis corresponding to Section~\ref{sec:causal_intervention}. The goal is to test whether the Fisher-IRG subspace has a functional role in downstream prediction, rather than merely exhibiting correlational separation between semantic-preserving and semantic-changing variation.

\vspace{-0.5em}
\paragraph{Intervention protocol.}
Let $U\in\mathbb{R}^{d\times k}$ denote an orthonormal basis of the Fisher-IRG subspace at the intervention layer. We decompose an anchor representation $z^{(a)}$ into
\[
z_{\mathrm{inv}}^{(a)}
=
UU^\top z^{(a)},
\qquad
z_{\mathrm{nuis}}^{(a)}
=
(I-UU^\top)z^{(a)}.
\]
For a donor representation $z^{(d)}$, we analogously obtain
$z_{\mathrm{inv}}^{(d)}$ and $z_{\mathrm{nuis}}^{(d)}$.
We then construct component-wise interventions by replacing one component of the anchor while holding the other fixed:
\[
\widetilde{z}_{\mathrm{inv}}
=
z_{\mathrm{nuis}}^{(a)}
+
z_{\mathrm{inv}}^{(d)},
\]
and
\[
\widetilde{z}_{\mathrm{nuis}}
=
z_{\mathrm{inv}}^{(a)}
+
z_{\mathrm{nuis}}^{(d)}.
\]
The intervened representation is propagated through the remaining frozen layers of the model, and the intervention effect is measured as the divergence between the original and intervened predictive outputs.

Donors are drawn either from the semantic-preserving set or the semantic-changing set of the same local semantic neighborhood. For donor type
$x\in\{\mathrm{SP},\mathrm{SC}\}$ and replaced component
$c\in\{\mathrm{inv},\mathrm{nuis}\}$, we denote the resulting output divergence by
$D_{x,c}$.

We use two complementary contrasts:
\[
\Delta_{\mathrm{causal}}
=
D_{\mathrm{SC},\mathrm{inv}}
-
D_{\mathrm{SP},\mathrm{inv}},
\]
and
\[
\Delta_{\mathrm{local}}
=
D_{\mathrm{SC},\mathrm{inv}}
-
D_{\mathrm{SC},\mathrm{nuis}}.
\]
The first tests \emph{semantic selectivity}: replacing the Fisher-IRG component with semantically different content should have a larger effect than replacing it with meaning-preserving content. The second tests \emph{geometric localization}: the same semantic change should exert a larger downstream effect when injected through the Fisher-IRG component than through its complementary component. Positive values for both contrasts therefore provide the expected causal-localization pattern.

All reported experiments use $k=16$ at layer 10. Language experiments use mean-pooled representations, whereas vision experiments use CLS-token representations.

\vspace{-0.5em}
\paragraph{Language-model interventions.}
To avoid selecting favorable examples separately for individual models, we evaluate the same three semantic groups, indexed by 1, 500, and 1000, across all seven language architectures. Table~\ref{tab:app_causal_language} reports the complete results.

\begin{table}[ht]
\centering
\scriptsize
\caption{Group-level causal intervention results for language models. $D_{\mathrm{SP},\mathrm{nuis}}$ and $D_{\mathrm{SC},\mathrm{nuis}}$ replace the complementary component using SP and SC donors, while $D_{\mathrm{SP},\mathrm{inv}}$ and $D_{\mathrm{SC},\mathrm{inv}}$ replace the Fisher-IRG component. Positive $\Delta_{\mathrm{causal}}$ and $\Delta_{\mathrm{local}}$ indicate semantic selectivity and localization, respectively. Results use mean-pooled representations at layer 10 with $k=16$.}
\label{tab:app_causal_language}
\setlength{\tabcolsep}{2.8pt}
\begin{tabular}{lccccccc}
\toprule
Model / Group
& $D_{\mathrm{SP,nuis}}\downarrow$
& $D_{\mathrm{SC,nuis}}\downarrow$
& $D_{\mathrm{SP,inv}}\downarrow$
& $D_{\mathrm{SC,inv}}\uparrow$
& $\Delta_{\mathrm{causal}}\uparrow$
& $\Delta_{\mathrm{local}}\uparrow$ \\
\midrule
Mistral-Instruct-v0.3 / 1
& .041859 & .069642 & .000323 & .084293 & .083970 & .014651 \\
Mistral-Instruct-v0.3 / 500
& .028741 & 1.214668 & .000060 & 3.002461 & 3.002401 & 1.787793 \\
Mistral-Instruct-v0.3 / 1000
& .018359 & 1.649377 & .000019 & 2.001373 & 2.001354 & .351996 \\
\midrule
LLaMA-3-Instruct / 1
& .033231 & 3.677786 & .000048 & 4.021666 & 4.021618 & .343880 \\
LLaMA-3-Instruct / 500
& .008527 & 1.752320 & .000030 & 1.915718 & 1.915688 & .163398 \\
LLaMA-3-Instruct / 1000
& .005825 & 1.000402 & .000194 & 1.010367 & 1.010173 & .009965 \\
\midrule
Gemma-IT / 1
& .090390 & 1.707979 & .000325 & 2.008799 & 2.008474 & .300820 \\
Gemma-IT / 500
& .042622 & .932285 & .001229 & 6.008258 & 6.007029 & 5.075973 \\
Gemma-IT / 1000
& .040759 & .151609 & .002384 & 2.008486 & 2.006102 & 1.856877 \\
\midrule
Qwen2.5-Instruct / 1
& .417850 & 6.319725 & .048583 & 9.492855 & 9.444272 & 3.173130 \\
Qwen2.5-Instruct / 500
& .050068 & 6.527780 & .001443 & 8.930468 & 8.929025 & 2.402688 \\
Qwen2.5-Instruct / 1000
& 1.379808 & 5.099013 & .006782 & 9.808725 & 9.801943 & 4.709712 \\
\midrule
DeepSeek-MoE-Chat / 1
& .031325 & .057812 & .000012 & .232485 & .232473 & .174673 \\
DeepSeek-MoE-Chat / 500
& .004810 & .437515 & .000085 & 1.015670 & 1.015585 & .578155 \\
DeepSeek-MoE-Chat / 1000
& .044327 & 1.487550 & .000141 & 4.024908 & 4.024767 & 2.537358 \\
\midrule
Phi-4-Mini / 1
& .018412 & .485730 & .000092 & 4.012745 & 4.012653 & 3.527015 \\
Phi-4-Mini / 500
& .036651 & 1.102610 & .000210 & 4.013874 & 4.013664 & 2.911264 \\
Phi-4-Mini / 1000
& .014725 & 3.494718 & .000064 & 7.048721 & 7.048657 & 3.554003 \\
\midrule
Falcon3-Instruct / 1
& .024450 & 3.761153 & .000117 & 5.014662 & 5.014545 & 1.253509 \\
Falcon3-Instruct / 500
& .043866 & 1.436206 & .000281 & 3.013217 & 3.012936 & 1.577011 \\
Falcon3-Instruct / 1000
& .015264 & .468874 & .000073 & .919986 & .919913 & .451112 \\
\bottomrule
\end{tabular}
\end{table}

All 21 model--group combinations exhibit positive
$\Delta_{\mathrm{causal}}$ and
$\Delta_{\mathrm{local}}$. In particular,
$D_{\mathrm{SP},\mathrm{inv}}$ remains small relative to
$D_{\mathrm{SC},\mathrm{inv}}$ throughout the table. Thus, the Fisher-IRG component is not generically sensitive to arbitrary replacement: substituting meaning-preserving content produces little predictive change, whereas substituting semantically changed content produces a substantially larger downstream response.

The localization contrast provides a second and distinct check. In every configuration,
$D_{\mathrm{SC},\mathrm{inv}}>D_{\mathrm{SC},\mathrm{nuis}}$.
Hence, semantic replacement is more consequential when introduced through the Fisher-IRG component than through its complement. The magnitude of this difference varies substantially across model families and semantic groups. For example, the localization margin for LLaMA-3-Instruct on group 1000 is relatively small, whereas several Gemma, Qwen, Phi, and DeepSeek configurations exhibit much larger margins. We therefore emphasize the consistent directional pattern rather than comparing absolute divergence magnitudes across architectures.

\vspace{-0.5em}
\paragraph{Vision interventions.}
We conduct the analogous analysis for ViT-B/16 using Visual Genome neighborhoods. To test whether the causal pattern depends on the type or difficulty of semantic change, we report independently selected examples under easy, medium, and hard SC conditions. Table~\ref{tab:app_causal_vision} gives the complete results.

\begin{table}[t]
\centering
\scriptsize
\caption{Group-level causal intervention results for ViT-B/16 across semantic-change difficulty. Results use CLS-token representations at layer 10 with $k=16$. Positive causal and localization gaps indicate that the Fisher-IRG component responds more strongly to semantic-changing than semantic-preserving replacement and more strongly than the complementary component under the same SC intervention.}
\label{tab:app_causal_vision}
\setlength{\tabcolsep}{3pt}
\begin{tabular}{lccccccc}
\toprule
SC / Group
& $D_{\mathrm{SP,nuis}}\downarrow$
& $D_{\mathrm{SC,nuis}}\downarrow$
& $D_{\mathrm{SP,inv}}\downarrow$
& $D_{\mathrm{SC,inv}}\uparrow$
& $\Delta_{\mathrm{causal}}\uparrow$
& $\Delta_{\mathrm{local}}\uparrow$ \\
\midrule
Easy / 300
& .386655 & .195308 & .003225 & .459543 & .456318 & .264235 \\
Easy / 550
& .114132 & .179609 & .001291 & .625532 & .624241 & .445923 \\
Easy / 800
& .231401 & .243617 & .003217 & .684134 & .680917 & .440517 \\
\midrule
Medium / 300
& .372400 & .037511 & .003132 & .107466 & .104334 & .069955 \\
Medium / 550
& .114486 & .212069 & .001107 & 1.006177 & 1.005070 & .794108 \\
Medium / 800
& .314213 & .114786 & .001499 & .741347 & .739848 & .626561 \\
\midrule
Hard / 300
& .142429 & .219710 & .001174 & .642299 & .641125 & .422589 \\
Hard / 550
& .114565 & .204601 & .001270 & .326446 & .325176 & .121845 \\
Hard / 800
& .098431 & .274156 & .003289 & .542316 & .539027 & .268160 \\
\bottomrule
\end{tabular}
\end{table}

The vision results reproduce both aspects of the language intervention pattern. First, $D_{\mathrm{SC},\mathrm{inv}}$ exceeds
$D_{\mathrm{SP},\mathrm{inv}}$ for all nine configurations, yielding positive causal gaps from $0.104334$ to $1.005070$. Second, semantic-changing replacement through Fisher-IRG produces a larger output change than replacement through the complementary component in every case, yielding positive localization gaps from $0.069955$ to $0.794108$.

The pattern persists across easy, medium, and hard semantic changes, but the magnitude is not monotonic in difficulty. We do not expect such monotonicity: the three conditions differ in the composition and type of semantic modification rather than constituting a calibrated scalar measure of semantic distance. The relevant observation is instead that the intervention asymmetry survives across all three regimes.

\vspace{-0.5em}
\paragraph{Relation to the geometric analyses.}
These interventions test a different property from the layer-wise selectivity and Grassmann experiments. Section~\ref{sec:selectivity_stability} establishes that Fisher-IRG provides stronger predictive semantic selectivity than Cov-IRG across depth, and also establishes that the recovered Fisher subspaces are generally more reproducible across independent estimates. Neither result alone demonstrates that manipulating the identified component changes downstream behavior in the predicted way.

The intervention experiment directly addresses this functional question. The positive causal gap shows that the Fisher-IRG component distinguishes SP from SC replacement, while the positive localization gap shows that the semantic effect is concentrated more strongly in Fisher-IRG than in its complementary component. Together, the two contrasts connect the geometric decomposition to model behavior.

\vspace{-0.5em}
\paragraph{Interpretation.}
The results should not be interpreted as implying a perfectly orthogonal semantic--nuisance decomposition. In several configurations, SC replacement of the complementary component also produces non-negligible output change, indicating that semantic information need not be exclusively confined to the Fisher-IRG span. The stronger and consistently selective response of the Fisher component instead supports a localization claim: predictive semantic sensitivity is concentrated in the Fisher-induced geometry relative to its complement.

Across seven language models and the evaluated vision setting, every reported intervention satisfies both the semantic-selectivity and localization criteria. This provides causal evidence that the Fisher-IRG subspace is functionally involved in semantically meaningful model behavior rather than merely reflecting correlational structure in the representation space.

%%%
%%%
%%%

%%%
%%%
%%% app validation

\vspace{-0.5em}
\section{Additional Results for Functional Validation}
\label{app:functional_validation}
\vspace{-0.5em}

This section provides the complete results corresponding to the two functional evaluations in Section~\ref{sec:functional_validation}: generalization of Fisher-IRG to held-out semantic perturbations and downstream semantic retrieval. In both experiments, the subspace is estimated exclusively from discovery variants, while evaluation uses disjoint held-out examples.

\paragraph{Held-out semantic separation.}
Given an orthonormal Fisher-IRG basis $U$, we project each representation as
\[
z_{\mathrm{proj}} = UU^\top z.
\]
For an anchor and its held-out semantic-preserving and semantic-changing variants, we measure the corresponding cosine distances $d_{\mathrm{SP}}$ and $d_{\mathrm{SC}}$ and define
\[
\Delta_{\mathrm{sep}}
=
d_{\mathrm{SC}}-d_{\mathrm{SP}}.
\]
A positive margin indicates that semantic changes are farther from the anchor than meaning-preserving perturbations. Reporting $d_{\mathrm{SP}}$ and $d_{\mathrm{SC}}$ separately further distinguishes whether an improvement arises from suppressing nuisance variation, amplifying semantic variation, or both.

For language models, Table~\ref{tab:app_semantic_sep_text} reports the complete results at three representative layers using mean-pooled representations and $k=16$. Fisher-IRG obtains the largest separation margin in all 21 model--layer comparisons.

\begin{table*}[t]
\centering
\scriptsize
\caption{Held-out semantic separation for language models. $d_{\mathrm{SP}}$ and $d_{\mathrm{SC}}$ denote cosine distances from the anchor to held-out semantic-preserving and semantic-changing variants, respectively, and $\Delta_{\mathrm{sep}}=d_{\mathrm{SC}}-d_{\mathrm{SP}}$. Subspaces are estimated from discovery variants and evaluated on disjoint held-out variants. Results use mean-pooled representations and $k=16$ and are averaged across semantic groups.}
\label{tab:app_semantic_sep_text}
\setlength{\tabcolsep}{2.7pt}
\begin{tabular}{lc|ccc|ccc|ccc}
\toprule
& &
\multicolumn{3}{c|}{Original}
& \multicolumn{3}{c|}{Fisher-IRG}
& \multicolumn{3}{c}{Cov-IRG} \\
Model & Layer
& $d_{\mathrm{SP}}\downarrow$
& $d_{\mathrm{SC}}\uparrow$
& $\Delta_{\mathrm{sep}}\uparrow$
& $d_{\mathrm{SP}}\downarrow$
& $d_{\mathrm{SC}}\uparrow$
& $\Delta_{\mathrm{sep}}\uparrow$
& $d_{\mathrm{SP}}\downarrow$
& $d_{\mathrm{SC}}\uparrow$
& $\Delta_{\mathrm{sep}}\uparrow$ \\
\midrule
Mistral-Instruct-v0.3
& 3  & .000019 & .000073 & .000053 & .000129 & .000563 & \textbf{.000434} & .000004 & .000019 & .000015 \\
& 8  & .000280 & .001212 & .000932 & .003044 & .015950 & \textbf{.012906} & .000062 & .000362 & .000300 \\
& 10 & .000421 & .001851 & .001430 & .005120 & .026828 & \textbf{.021707} & .000094 & .000568 & .000474 \\
\midrule
Llama-3-Instruct
& 3  & .000058 & .000196 & .000137 & .000175 & .000743 & \textbf{.000568} & .000015 & .000059 & .000043 \\
& 8  & .000368 & .001709 & .001341 & .000788 & .004143 & \textbf{.003356} & .000081 & .000497 & .000416 \\
& 10 & .000455 & .002103 & .001648 & .000911 & .005168 & \textbf{.004257} & .000103 & .000605 & .000502 \\
\midrule
Gemma-IT
& 3  & .000126 & .000866 & .000740 & .002274 & .007521 & \textbf{.005247} & .000053 & .000099 & .000046 \\
& 8  & .000154 & .001280 & .001126 & .000164 & .003617 & \textbf{.003453} & .000114 & .000141 & .000027 \\
& 10 & .000233 & .001367 & .001134 & .003310 & .012246 & \textbf{.008936} & .000068 & .000343 & .000275 \\
\midrule
Qwen2.5-Instruct
& 3  & .000147 & .000720 & .000573 & .003551 & .010712 & \textbf{.007161} & .000148 & .000225 & .000077 \\
& 8  & .000271 & .001475 & .001204 & .002814 & .008175 & \textbf{.005361} & .000104 & .000213 & .000109 \\
& 10 & .000175 & .000748 & .000573 & .000273 & .005617 & \textbf{.005344} & .000076 & .000128 & .000052 \\
\midrule
DeepSeek-MoE-Chat
& 3  & .000407 & .001428 & .001021 & .003616 & .009168 & \textbf{.005552} & .000082 & .000264 & .000182 \\
& 8  & .000094 & .000973 & .000879 & .002615 & .003600 & \textbf{.000985} & .000073 & .000303 & .000230 \\
& 10 & .000400 & .001343 & .000943 & .002253 & .003794 & \textbf{.001541} & .000113 & .000253 & .000140 \\
\midrule
Phi-4-Mini
& 3  & .000218 & .001171 & .000953 & .004090 & .007564 & \textbf{.003474} & .000036 & .000267 & .000231 \\
& 8  & .000325 & .001364 & .001039 & .003642 & .007919 & \textbf{.004277} & .000078 & .000082 & .000004 \\
& 10 & .000246 & .000894 & .000648 & .000420 & .015553 & \textbf{.015133} & .000059 & .000102 & .000043 \\
\midrule
Falcon3-Instruct
& 3  & .000375 & .001170 & .000795 & .000756 & .012498 & \textbf{.011742} & .000146 & .000316 & .000170 \\
& 8  & .000168 & .001196 & .001028 & .003203 & .004681 & \textbf{.001478} & .000107 & .000247 & .000140 \\
& 10 & .000506 & .001331 & .000825 & .004043 & .008521 & \textbf{.004478} & .000044 & .000063 & .000019 \\
\bottomrule
\end{tabular}
\end{table*}

The improvement is broad across architectures and depth, but its source varies by model and layer. In many language settings, Fisher-IRG primarily increases $d_{\mathrm{SC}}$, indicating stronger sensitivity to semantic displacement, rather than merely shrinking SP distances. The consistent improvement in $\Delta_{\mathrm{sep}}$ nevertheless shows that the semantic-versus-nuisance structure recovered from the discovery variants transfers to unseen perturbations. The complete language values are given in Table~\ref{tab:app_semantic_sep_text}. 

\paragraph{Held-out visual separation.}
For ViT-B/16, each Visual Genome group uses seven SP and 50 SC variants for subspace discovery, with the remaining three SP and 14 SC variants reserved for held-out evaluation. The discovery SC variants are pooled for subspace estimation, while easy, medium, and hard semantic-change conditions are retained separately for downstream evaluation. All results use CLS-token representations with $k=16$.

\begin{table*}[t]
\centering
\scriptsize
\caption{Held-out semantic separation for ViT-B/16 across semantic-change difficulty and representative layers. Fisher-IRG achieves the largest separation margin in every reported configuration.}
\label{tab:app_semantic_sep_vision}
\setlength{\tabcolsep}{2.8pt}
\begin{tabular}{lc|ccc|ccc|ccc}
\toprule
& &
\multicolumn{3}{c|}{Original}
& \multicolumn{3}{c|}{Fisher-IRG}
& \multicolumn{3}{c}{Cov-IRG} \\
SC Difficulty & Layer
& $d_{\mathrm{SP}}\downarrow$
& $d_{\mathrm{SC}}\uparrow$
& $\Delta_{\mathrm{sep}}\uparrow$
& $d_{\mathrm{SP}}\downarrow$
& $d_{\mathrm{SC}}\uparrow$
& $\Delta_{\mathrm{sep}}\uparrow$
& $d_{\mathrm{SP}}\downarrow$
& $d_{\mathrm{SC}}\uparrow$
& $\Delta_{\mathrm{sep}}\uparrow$ \\
\midrule
Easy
& 3  & .004884 & .003291 & -.001592 & .007024 & .033447 & \textbf{.026423} & .006320 & .003236 & -.003084 \\
& 8  & .032660 & .054270 & .021609 & .022415 & .551965 & \textbf{.529550} & .042919 & .072215 & .029296 \\
& 10 & .121349 & .229988 & .108639 & .040845 & .764698 & \textbf{.723853} & .070417 & .213771 & .143354 \\
\midrule
Medium
& 3  & .004889 & .004332 & -.000556 & .007745 & .043607 & \textbf{.035862} & .005922 & .004327 & -.001595 \\
& 8  & .032724 & .068977 & .036253 & .017568 & .595776 & \textbf{.578208} & .039122 & .090228 & .051106 \\
& 10 & .121571 & .289566 & .167995 & .027675 & .752244 & \textbf{.724569} & .066331 & .271418 & .205087 \\
\midrule
Hard
& 3  & .004859 & .004471 & -.000389 & .007749 & .042007 & \textbf{.034258} & .005797 & .004445 & -.001352 \\
& 8  & .032585 & .064877 & .032292 & .018366 & .547136 & \textbf{.528770} & .040094 & .086060 & .045966 \\
& 10 & .121252 & .270515 & .149263 & .029561 & .707937 & \textbf{.678376} & .066392 & .242331 & .175939 \\
\bottomrule
\end{tabular}
\end{table*}

The vision results reveal a particularly clear depth-dependent transition. At layer 3, both the original representation and Cov-IRG exhibit negative separation margins for all three difficulty levels, whereas Fisher-IRG already produces positive margins. By layer 10, Fisher-IRG reaches margins of $0.723853$, $0.724569$, and $0.678376$ for easy, medium, and hard changes, substantially exceeding both comparison representations. At the later layers, the improvement reflects both reduced SP distance and substantially increased SC distance, indicating simultaneous nuisance suppression and semantic amplification. 

\begin{table}[h]
\centering
\small
\caption{Held-out semantic retrieval across language models. Results use mean-pooled representations at layer 10 with $k=16$ and are averaged over 1,800 semantic groups. Bold indicates the best result for each model and metric.}
\label{tab:app_semantic_retrieval}
\setlength{\tabcolsep}{4pt}
\begin{tabular}{llccc}
\toprule
Model & Representation
& Pairwise $\uparrow$
& MRR $\uparrow$
& Hit@1 $\uparrow$ \\
\midrule
Mistral-Instruct-v0.3
& Original   & .974 & .961 & .942 \\
& Random     & .965 & .921 & .850 \\
& PCA        & .961 & .934 & .889 \\
& cPCA       & .979 & .959 & .928 \\
& Cov-IRG    & .957 & .902 & .810 \\
& Fisher-IRG & \textbf{.985} & \textbf{.975} & \textbf{.957} \\
\midrule
LLaMA-3-Instruct
& Original   & \textbf{.978} & .952 & .910 \\
& Random     & .954 & .913 & .845 \\
& PCA        & .961 & .941 & .904 \\
& cPCA       & .972 & .953 & .923 \\
& Cov-IRG    & .949 & .885 & .780 \\
& Fisher-IRG & .968 & \textbf{.956} & \textbf{.933} \\
\midrule
Gemma-IT
& Original   & .966 & .943 & .905 \\
& Random     & .957 & .923 & .865 \\
& PCA        & .932 & .893 & .821 \\
& cPCA       & .954 & .895 & .799 \\
& Cov-IRG    & .949 & .876 & .755 \\
& Fisher-IRG & \textbf{.979} & \textbf{.959} & \textbf{.925} \\
\midrule
Qwen2.5-Instruct
& Original   & .912 & .873 & .815 \\
& Random     & .857 & .805 & .710 \\
& PCA        & .927 & .871 & .784 \\
& cPCA       & .911 & .864 & .801 \\
& Cov-IRG    & .886 & .869 & \textbf{.825} \\
& Fisher-IRG & \textbf{.931} & \textbf{.880} & .807 \\
\midrule
DeepSeek-MoE-Chat
& Original   & .945 & .897 & .818 \\
& Random     & .915 & .864 & .787 \\
& PCA        & .938 & .886 & .804 \\
& cPCA       & .945 & .898 & .819 \\
& Cov-IRG    & .923 & .871 & .792 \\
& Fisher-IRG & \textbf{.951} & \textbf{.904} & \textbf{.837} \\
\midrule
Phi-4-Mini
& Original   & .943 & .922 & .876 \\
& Random     & .872 & .839 & .755 \\
& PCA        & .914 & .861 & .770 \\
& cPCA       & .932 & .879 & .798 \\
& Cov-IRG    & .859 & .827 & .739 \\
& Fisher-IRG & \textbf{.956} & \textbf{.929} & \textbf{.881} \\
\midrule
Falcon3-Instruct
& Original   & .924 & .896 & .843 \\
& Random     & .889 & .867 & .811 \\
& PCA        & .911 & .886 & .831 \\
& cPCA       & .919 & .889 & .817 \\
& Cov-IRG    & .872 & .871 & .839 \\
& Fisher-IRG & \textbf{.941} & \textbf{.914} & \textbf{.856} \\
\bottomrule
\end{tabular}
\end{table}

\paragraph{Held-out semantic retrieval.}
We next evaluate whether the geometric separation above transfers to a direct semantic matching task. For each of the 1,800 text semantic groups, the original prompt serves as the query, one held-out SP variant is the relevant candidate, and five held-out SC variants are semantic distractors. None of the retrieval candidates is used to estimate the subspace.

\vspace{-1.0em}
\paragraph{Contrastive PCA baseline.}
Contrastive PCA (cPCA) identifies directions whose variance is enriched
in a target population relative to a background population. For each
anchor, we treat the discovery SC variants as the target and the
discovery SP variants as the background. Let $C_{\mathrm{sc}}$ and
$C_{\mathrm{sp}}$ denote their empirical representation covariance
matrices. Given a contrast parameter $\alpha \geq 0$, cPCA forms
\[
C_{\alpha}
=
C_{\mathrm{sc}}-\alpha C_{\mathrm{sp}}
\]
and selects directions that maximize
\[
\max_{\|v\|_2=1}
v^\top C_{\alpha}v
=
\max_{\|v\|_2=1}
\left(
v^\top C_{\mathrm{sc}}v
-
\alpha v^\top C_{\mathrm{sp}}v
\right).
\]
We construct $U_{\mathrm{cPCA}}$ from the orthonormal eigenvectors
associated with the $k$ largest algebraic eigenvalues of $C_{\alpha}$.
Thus, cPCA uses a weighted difference of semantic and nuisance
variances, whereas \cn{} uses their regularized generalized Rayleigh
quotient. The resulting cPCA subspace is evaluated using the same
projection and held-out retrieval protocol as the other baselines.

Both query and candidate representations are projected using
\[
z_{\mathrm{proj}}=UU^\top z
\]
and ranked by cosine distance. 
We compare Fisher-IRG with the original unprojected representation and dimension-matched Random, PCA, cPCA, and Cov-IRG subspaces. 
SP--SC Pairwise Accuracy measures how often the held-out SP candidate is closer than an SC distractor; MRR measures the reciprocal rank of the SP candidate; and Hit@1 records whether it is ranked first.

Fisher-IRG achieves the highest MRR for all seven architectures and the highest pairwise accuracy and Hit@1 for six of seven. The exceptions are informative rather than contradictory. For Llama-3-Instruct, the original representation has slightly higher pairwise accuracy ($0.978$ versus $0.968$), while Fisher-IRG improves both MRR ($0.956$ versus $0.952$) and Hit@1 ($0.933$ versus $0.910$). For Qwen2.5-Instruct, Fisher-IRG achieves the highest pairwise accuracy and MRR, while Cov-IRG has a slightly higher Hit@1 ($0.825$ versus $0.807$). 

The additional baselines test whether retrieval performance can be
explained by random dimensionality reduction or variance-based subspace
selection. \mn{} achieves higher MRR than Random, PCA, cPCA, and \cn{}
for every evaluated architecture. Its pairwise accuracy also exceeds
these baselines except for cPCA on LLaMA-3-Instruct
(0.968 versus 0.972). Full per-model results are reported in Table~\ref{tab:app_semantic_retrieval}.

\paragraph{Summary.}
The two evaluations test complementary consequences of the recovered geometry. Held-out semantic separation asks whether the SP/SC organization discovered locally persists on unseen perturbations; retrieval asks whether that organization can be exploited for direct semantic matching. Fisher-IRG consistently improves the former across language and vision and generally improves the latter across language architectures. Together, these results support the conclusion that the Fisher-induced subspace generalizes beyond its discovery samples and retains functionally useful semantic structure.

%%%
%%%
%%%

%%%
%%%
%%% implement

\section{Implementation Details}
\label{app:implementation}

\paragraph{Text experiments.}
We implement the language experiments in PyTorch using Hugging Face Transformers. We evaluate Mistral-Instruct-v0.3, Llama-3-Instruct, Gemma-IT, Qwen2.5-Instruct, DeepSeek-MoE-Chat, Phi-4-Mini, and Falcon3-Instruct, with all model parameters frozen and each model run in evaluation mode. Inputs are processed using the corresponding tokenizer and truncated to a maximum length of 128 tokens. Unless otherwise stated, layer representations are obtained by mean-pooling the hidden states of all non-padding tokens.

For interventions on a mean-pooled representation, a displacement is lifted back to token space by adding the same displacement to every non-padding token state at the selected layer. The modified token states are then propagated through the remaining transformer blocks and final normalization, and next-token probabilities at the final valid prompt position define the predictive readout.

For \mn, we compute gradients of next-token log probabilities with respect to an additive perturbation at the selected hidden layer. At each representation $z$, we retain the top-$q$ outcomes under the local predictive distribution $p(\cdot\mid z)$ and renormalize their probabilities over the retained set. The resulting truncated Fisher estimator is
\[
\widehat F_q(z)
=
\sum_{y\in T_q(z)}
\frac{p(y\mid z)}{m_q(z)}
\,g_y(z)g_y(z)^\top,
\qquad
m_q(z)
=
\sum_{y\in T_q(z)}p(y\mid z),
\]
where $g_y(z)=\nabla_z\log p(y\mid z)$. Thus, both the retained outcomes and their weights are determined locally at the representation being analyzed, consistent with the Fisher approximation in Section~\ref{sec:fisher_discovery}. The score gradients are computed from the original predictive distribution; truncation and renormalization are used only to approximate the Fisher expectation over the output space.

Both Fisher and covariance operators are represented in low-rank factored form, avoiding explicit construction of dense hidden-dimension-by-hidden-dimension matrices. The generalized eigenproblem is solved within the joint span of the SP and SC factors. For numerical stability, the SP operator is regularized before whitening by adding
\(
10^{-3}
\)
times its mean diagonal value to the diagonal. The main experiments use subspace dimension $k=16$, with $k\in\{8,32\}$ included in the dimension analysis. The pooling ablation compares mean pooling with last-token pooling while holding the model, layer, semantic groups, and subspace dimension fixed. Predictive intervention effects are measured at the final token using KL divergence; cosine similarity and mean-squared error are used as supplementary representation-level measures where reported.

For every reported configuration, we restrict the retained subspace dimension to the effective rank supported by the corresponding generalized-eigenvalue problem. We retain only directions associated with numerically nonzero generalized eigenvalues and never pad a recovered basis with null directions. The discovery-set sizes used in the reported experiments satisfy this condition for each evaluated $k$.

\paragraph{Vision experiments.}
We evaluate pretrained ViT-B/16, DINOv2 ViT-B/14, and CLIP ViT-B/16 encoders with all model parameters frozen. Images are processed using the standard input resolution and normalization associated with each pretrained model, and layer representations are extracted from the CLS token. 
Each Visual Genome neighborhood contains one anchor, 10 semantic-preserving variants and 64 semantic-changing variants spanning easy, medium, and hard conditions.

A representation displacement is applied to every token state, including the CLS token, immediately before transformer block $\ell$. The modified token sequence is propagated through block $\ell$ and all remaining blocks, followed by the model's final normalization. For ViT-B/16, the fixed classifier applied to the resulting CLS representation defines the predictive distribution. For DINOv2 ViT-B/14, we use a linear classification readout that is fitted before geometric analysis and then held fixed during Fisher estimation and all subsequent interventions. For CLIP ViT-B/16, image representations are scored against a fixed set of candidate text representations using cosine similarity; the candidate set is held fixed throughout each analysis. The resulting vision scores are converted to predictive probabilities using a softmax temperature of $\tau=0.5$.

For each semantic group, the vision subspaces are estimated from the pooled discovery SC set, while easy, medium, and hard semantic-change conditions are retained separately for downstream evaluation.
We use the same low-rank generalized-eigenvalue solver and relative regularization coefficient of $10^{-3}$ as in the language experiments. The main vision experiments use $k=16$, with $k\in\{8,32\}$ included in the subspace-dimension analysis. Results are first computed within each Visual Genome neighborhood and then aggregated across groups, with the three SC difficulty levels reported separately where applicable.

\paragraph{Subspace stability and Grassmann evaluation.}
To evaluate whether the recovered geometry is reproducible under variation in the sampled local neighborhood, we independently bootstrap the SP and SC variants within each semantic group. Each bootstrap replicate samples, with replacement, the same number of SP and SC variants as the original neighborhood and reconstructs the corresponding subspaces for \mn, \cn, PCA, and cPCA. We use random seed 42 and three bootstrap replicates per group--layer configuration, yielding the three pairwise subspace comparisons among replicates.

For two orthonormal bases $\mathbf{U}_1$ and $\mathbf{U}_2$, the singular values of
\(
\mathbf{U}_1^\top \mathbf{U}_2, 
\)
define the principal cosines. Writing the associated principal angles as
\(
\boldsymbol{\theta}
=
(\theta_1,\ldots,\theta_k),
\)
we compute the Grassmann geodesic distance as
\(
d_{\mathrm{Gr}}(\mathbf{U}_1,\mathbf{U}_2)
=
\left\|\boldsymbol{\theta}\right\|_2.
\)
We additionally report normalized projection distance, mean principal cosine, and the absolute cosine similarity between corresponding individual basis vectors. Higher principal cosine and lower Grassmann or projection distance indicate greater subspace reproducibility. Comparing subspace-level measures with individual-vector cosine further distinguishes stability of the recovered subspace from stability of a particular basis representation. A random orthonormal subspace of the same dimension is included as a reference baseline.

The reported stability experiments use the same sample-local Fisher estimator described above with $q=8$, mean-pooled language representations, and $k\in\{8,16,32\}$.  
For vision, the corresponding CLS-token representations are used. SP and SC variants are resampled independently so that stability reflects sensitivity to the finite local neighborhood rather than repeated evaluation of the same sample set.

\paragraph{Computational Resources.}
All experiments were conducted on workstations equipped with NVIDIA GeForce RTX 4090 GPUs, each with 24\,GB of memory. Model forward passes were performed in half precision, while pooled representations, Fisher gradients, low-rank operators, eigendecompositions, and evaluation metrics were stored or computed in single precision where needed for numerical stability. Sentence- and image-level representations were cached on CPU and transferred to the GPU only for the layer and operator currently being evaluated. This implementation avoids constructing or retaining dense Fisher matrices and keeps the memory footprint compatible with the available hardware.

%%%
%%%
%%%

\end{document}